\documentclass[a4paper, 10pt, conference]{ieeeconf}      
\usepackage{FG2025}

\FGfinalcopy 

\IEEEoverridecommandlockouts                              
\usepackage{amsmath} 
\usepackage{amssymb}  
\usepackage{graphicx}
\usepackage{booktabs}
\usepackage{multirow}
\usepackage{listings}
\usepackage{bm}

\usepackage{hyperref}

\def\FGPaperID{244} 

\title{\LARGE \bf
ID-Booth: Identity-consistent Face Generation with Diffusion Models
}

\author{\parbox{17cm}{\centering
    {\large Darian Tomašević$^1$, Fadi Boutros$^2$, Chenhao Lin$^3$, Naser Damer$^{2,4}$, Vitomir Štruc$^5$ and Peter Peer$^1$}\\
    {\normalsize
    $^1$ University of Ljubljana, Faculty of Computer and Information Science, Ljubljana, Slovenia\\
    $^2$ Fraunhofer Institute for Computer Graphics Research IGD, Darmstadt, Germany\\
    $^3$ Xi’an Jiaotong University, School of Cyber Science and Engineering, Xi’an, China\\ 
    $^4$ Department of Computer Science, TU Darmstadt, Germany \\
    $^5$ University of Ljubljana, Faculty of Electrical Engineering, Ljubljana, Slovenia
    }}
    \thanks{Supported in parts by the Slovenian Research and Innovation Agency (ARIS) through Research Programmes P2-0250 (B) "Metrology and Biometric Systems" and P2--0214 (A) “Computer Vision”, the ARIS Project J2-50065 "DeepFake DAD", and the ARIS Young Researcher Programme.
    }
}

\usepackage{fancyhdr}
\begin{document}

\ifFGfinal
\thispagestyle{empty}
\pagestyle{empty}
\else
\author{Anonymous FG2025 submission\\ Paper ID \FGPaperID \\}
\pagestyle{plain}
\fi
\maketitle

\thispagestyle{fancy}

\begin{abstract} 


Recent advances in generative modeling have enabled the generation of high-quality synthetic data that is applicable in a variety of domains, including face recognition. Here, state-of-the-art generative models typically rely on conditioning and fine-tuning of powerful pretrained diffusion models to facilitate the synthesis of realistic images of a desired identity. Yet, these models often do not consider the identity of subjects during training, leading to poor consistency between generated and intended identities. In contrast, methods that employ identity-based training objectives tend to overfit on various aspects of the identity, and in turn, lower the diversity of images that can be generated. To address these issues, we present in this paper a novel generative diffusion-based framework, called ID-Booth. ID-Booth consists of a denoising network responsible for data generation, a variational auto-encoder for mapping images to and from a lower-dimensional latent space and a text encoder that allows for prompt-based control over the generation procedure. The framework utilizes a novel triplet identity training objective and enables identity-consistent image generation while retaining the synthesis capabilities of pretrained diffusion models. Experiments with a state-of-the-art latent diffusion model and diverse prompts reveal that our method facilitates better intra-identity consistency and inter-identity separability than competing methods, while achieving higher image diversity. In turn, the produced data allows for effective augmentation of small-scale datasets and training of better-performing recognition models in a privacy-preserving manner. The source code for the ID-Booth framework is publicly available at \url{https://github.com/dariant/ID-Booth}. 
\end{abstract}


\section{Introduction}
Deep learning models are nowadays utilized as backbones in a variety of recognition systems~\cite{bai2021explainable}. These models typically require sufficiently large and diverse training datasets to achieve competitive performance. However, obtaining suitable datasets can be difficult in the field of biometrics, due to copyright, consent, and privacy issues~\cite{jasserand2018massive, meden2021privacy}. With recent advancements in generative models, researchers are increasingly exploring the use of synthetic data to address the data needs of contemporary deep learning models. This is especially true in face recognition, where synthetic data may be used to train models or augment existing datasets by enriching the variation present in real-world data~\cite{boutros2023synthetic}.  


\begin{figure}[tb!]
    \centering
    \includegraphics[width=\linewidth]{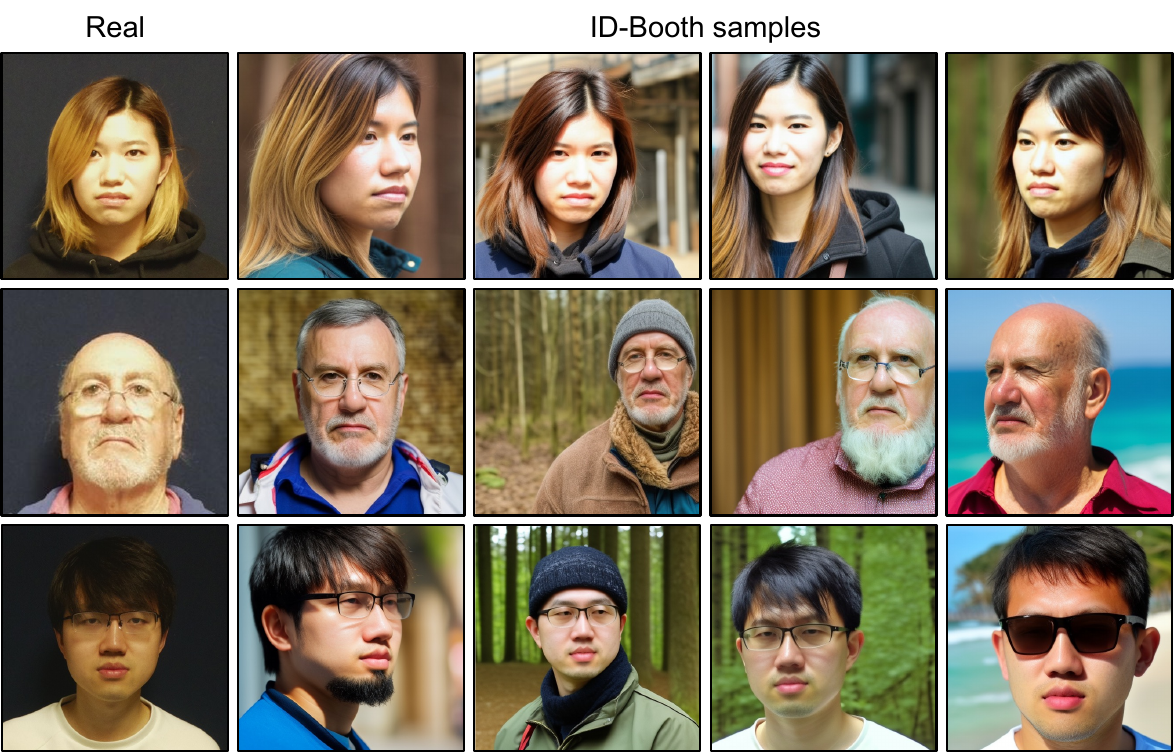}
    \vspace{-6mm}
    \caption{\textbf{Samples generated with the proposed ID-Booth framework.} The framework enables fine-tuning of pretrained diffusion models for generating diverse identity-consistent face images based on images gathered in a constrained setting with the consent of subjects. 
    } 
    \label{fig:IDBooth_diverse_samples}
    \vspace{-6mm}
\end{figure}

State-of-the-art generative models are currently dominated by diffusion-based techniques, which offer unparalleled synthesis capabilities in terms of quality and diversity of the generated data, while enabling synthesis guided by text prompts~\cite{rombach2022high_latent_diffusion}. 
Recently, diffusion models have also been utilized to produce datasets suitable for face recognition tasks, i.e., containing images of multiple identities with multiple samples each. 
To this end, approaches rely on identity-conditioning~\cite{papantoniou2024arc2face,wang2024instantid,ye2023ip_adapter} and fine-tuning~\cite{ruiz2023dreambooth} of pretrained diffusion models.  
Nevertheless, most solutions focus mainly on image reconstruction during training, resulting in poor consistency between the desired and the generated identities. 
To address this issue, PortraitBooth~\cite{peng2024portraitbooth} proposed an additional identity-based training objective, which can be used to extend the fine-tuning DreamBooth~\cite{ruiz2023dreambooth} method. However, the solution only considers the identity similarity of input samples and the generated samples during training. In turn, it tends to overfit on input identity features, including undesired characteristics, e.g., the pose, age, hair, accessories, thereby reducing the diversity of generated images.

In this paper, we present a solution for the outlined issues, in the form of a new generative framework, called ID-Booth. The proposed framework entails three main components, including $(i)$ a denoising network that produces data based on input noise, $(ii)$ a Variational Auto-Encoder (VAE) that maps images to and from a more efficient latent space on which the denoising network operates, and $(iii)$ a text encoder that enables prompt-based conditioning of the denoising network.
The proposed framework utilizes a novel triplet identity objective, which considers both positive and negative identity samples during training, to facilitate the generation of identity-consistent images while retaining the synthesis capabilities of pretrained models. 
Throughout the experiments, we explore the suitability of ID-Booth for addressing privacy concerns by generating diverse synthetic in-the-wild images of identities from the Tufts Face Database~\cite{TUFTS_panetta2018comprehensive}, which contains images gathered in a constrained setting with subject consent, as shown in Figure~\ref{fig:IDBooth_diverse_samples}.
We perform fine-tuning of a state-of-the-art diffusion model conditioned on diverse prompts and compare synthesis results with  DreamBooth~\cite{ruiz2023dreambooth} and a PortraitBooth-based~\cite{peng2024portraitbooth} version of it in terms of image quality, fidelity and diversity as well as intra-identity consistency and inter-identity separability. Furthermore, we investigate the real-world utility of the produced synthetic samples for augmenting existing datasets to train modern face recognition models in a privacy-preserving manner. We demonstrate that our fine-tuning framework enables the generation of more diverse synthetic samples with better intra-identity consistency and inter-identity separability. As showcased by improved recognition performance, across five real-world verification benchmarks, this makes our approach more suitable for augmenting small-scale training datasets than existing solutions~\cite{ruiz2023dreambooth,peng2024portraitbooth}. 
Overall, the paper makes the following contributions: 
\begin{itemize}
    \item We introduce ID-Booth, a novel  framework for generating highly-diverse identity-consistent privacy-preserving face images.  
    \item We propose a novel triplet identity learning objective for fine-tuning that improves identity consistency while retaining better image diversity. 
    \item We demonstrate the suitability of the produced data for augmenting existing small-scale datasets and show that training with the mixed images leads to better performing face recognition models. 
\end{itemize}

\section{Related work} 
\noindent\textbf{Image generation.}~The field of image synthesis has undergone rapid development since the introduction of deep generative models. Generative Adversarial Networks (GANs)~\cite{goodfellow2014generative} were the initial models to achieve the synthesis of convincing images, with a generator and a discriminator network. Extensive improvements followed, namely StyleGAN~\cite{stylegan_1_karras2019style} facilitated higher image quality and better control over the generation process. However, the synthesis capabilities of GANs have nowadays been surpassed by recent diffusion models~\cite{dhariwal2021diffusion_vs_gan}, which generate images by gradually removing noise from initial noisy samples. This denoising process is learned with a convolutional encoder-decoder by predicting the noise that is added to training samples at different scales~\cite{ho2020denoising}. Recently, Latent Diffusion Models (LDMs)~\cite{rombach2022high_latent_diffusion} achieved improved  efficiency and efficacy  by moving the denoising process from the pixel space to a lower-dimensionality latent space of a pretrained variational autoencoder. Their remarkable synthesis capabilities and conditioning on text prompts via a pretrained text encoder have led to their broad adoption, namely of the open-source Stable Diffusion (SD) model~\cite{rombach2022high_latent_diffusion}. 
Image resolution of these models has been further improved by utilizing a larger U-Net backbone along with two text encoders and additional conditioning schemes~\cite{podell2024sdxl}. Recent approaches have also enhanced control over the generation process, e.g., ControlNet~\cite{zhang2023controlnet} conditions the model on segmentation masks or depth maps via an auxiliary trainable copy of the model, while IP-Adapter~\cite{ye2023ip_adapter} utilizes image features as a condition through a decoupled cross-attention mechanism. Fine-tuning approaches have also been developed to incorporate new concepts into pretrained diffusion models by training on a minimal set of images~\cite{ruiz2023dreambooth}.


\vspace{0.9mm}\noindent\textbf{Generating synthetic face recognition data.} 
Generative models and synthetic data hold considerable potential in face recognition by enabling the creation of large-scale (training and test) datasets with predefined characteristics, facilitating augmentation in data-scarce application scenarios, and balancing data across different demographics~\cite{boutros2023synthetic}. To enable control over various characteristics of generated faces, Deng~\textit{et al.}~\cite{deng2020disentangled} conditioned  StyleGAN~\cite{stylegan_1_karras2019style} on input 3D face priors.
However, recognition models trained on the generated data achieved worse performance than those trained on real-world data. To tackle this issue, Qiu \textit{et al.}~\cite{qiu2021synface} introduced identity and domain mixup of synthetic and real data during training.
Boutros \textit{et al.}~\cite{boutros2022sface} proposed to condition StyleGAN2~\cite{stylegan2_ADA_karras2020training} on one-hot encoded identity labels. This improved intra-identity diversity at the cost of lowered inter-identity separability and a limited amount of possible identities. To address this, Tomašević \textit{et al.}~\cite{tomasevic2024arcbifacegan} instead utilized identity features from a pretrained face recognition model as the condition, in addition to enabling the generation of multispectral data.

Recently, Boutros \textit{et al.}~\cite{boutros2023idiff} achieved the generation of identity-specific images with latent diffusion models by conditioning the denoising network on face recognition features. The proposed contextual partial dropout also prevented overfitting on identities and enabled control over inter-identity separability and intra-identity diversity. 
Differently, more recent approaches relied on pretrained diffusion models~\cite{rombach2022high_latent_diffusion} rather than training the models from scratch.
Ruiz \textit{et al.}~\cite{ruiz2023dreambooth} presented the DreamBooth method that 
can associate a new identity to a rare text token through fine-tuning on images of the identity. During training, face images generated by the pretrained model are also used to preserve prior synthesis capabilities.
Arc2Face~\cite{papantoniou2024arc2face} instead replaces the identity token with recognition features and fine-tunes the model on a large-scale dataset. The textual-part of the prompt is frozen, so that control is tied primarily to the identity features, thus enabling more consistent generation of input identities. However, this comes at the cost of losing powerful prompt-based control. 
The recent IP-Adapter~\cite{ye2023ip_adapter} has also been modified to use identity features as the condition, while retaining control of text prompts  through decoupled cross-attention.  
InstantID~\cite{wang2024instantid} extends these capabilities by incorporating spatial control with an auxiliary ControlNet-based~\cite{zhang2023controlnet} module conditioned on facial landmarks and features. 
Despite advancements, identity consistency remained problematic, as the identity aspect was not considered in training objectives. 
To address this, Peng \textit{et al.}~\cite{peng2024portraitbooth} introduced PortraitBooth, a method that incorporates an identity-based objective into the training process, which can also be applied to the fine-tuning of DreamBooth~\cite{ruiz2023dreambooth}. However, the solution only relies on the identity similarity of training images and generated noisy images, despite the success of more refined objectives on face recognition tasks~\cite{trigueros2018enhancing}. As a result, the approach can overfit even on undesired characteristics of training identities, e.g., their pose, age or face accessories, which lowers the diversity of produced images. 
In contrast, our proposed ID-Booth framework utilizes a triplet objective that relies on the identity similarity between generated images and both training images (i.e., positive samples) and prior images produced by the initial model (i.e., negative samples). This enables better identity consistency, while better retaining synthesis capabilities of pretrained latent diffusion models.

\begin{figure*}[t!]
    \centering
    \includegraphics[width=0.9\linewidth]{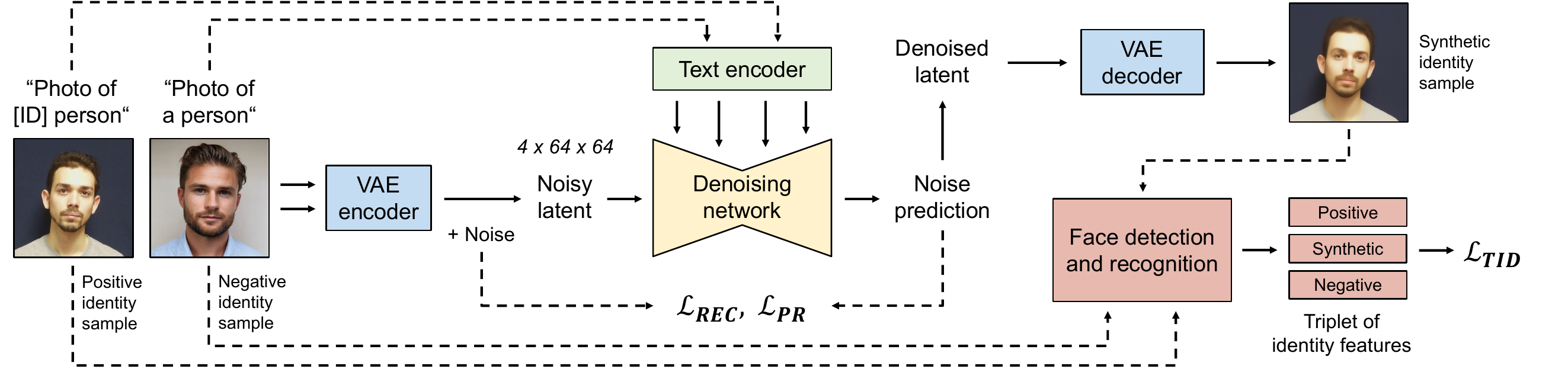}
    \caption{\textbf{Overview of the ID-Booth framework.} The framework utilizes three training objectives to fine-tune a pretrained diffusion model. $\mathcal{L}_{REC}$ and $\mathcal{L}_{PR}$ are aimed at the reconstruction of training and prior images. Differently, the proposed triplet identity objective $\mathcal{L}_{TID}$ focuses on the identity similarity between generated samples and both training and prior samples to improve identity consistency without impacting the capabilities of the pretrained model.
    \vspace{-6mm}
    } 
    \label{fig:ID-Booth_framework}
\end{figure*}

\section{Methodology}
In this section we present the inner workings of ID-Booth, a framework for generating diverse high-fidelity identity-consistent facial images suitable for augmenting small-scale datasets captured with the consent of subjects. 


\subsection{The ID-Booth framework}

The proposed diffusion-based ID-Booth framework consists of three primary components, as depicted in Figure~\ref{fig:ID-Booth_framework}. This includes $(i)$ the denoising network, responsible for enabling data generation through diffusion, $(ii)$ the Variational Auto-Encoder (VAE) that maps images to and from a more efficient latent space, and $(iii)$ the text encoder, which enables prompt-based control of the generation process.
Fine-tuning of pretrained diffusion models is then achieved with three training objectives, $(i)$ the conventional reconstruction loss on a small set of input samples, $(ii)$ the prior preservation loss, focused on combating overfitting via the reconstruction of images generated before fine-tuning, and $(iii)$ the triplet identity loss, which utilizes a pretrained face recognition model to guide the diffusion model toward better similarity between generated and target identities rather than random identities from prior images. Details of each component and objective are provided below. 
 


\vspace{0.8mm}\noindent\textbf{Denoising network.}~At the core of the diffusion model lies the denoising network, which is trained to reverse a noising process $q$ that gradually degrades training images by adding noise at different scales. 
This entails the corruption of a sample $x_0$ from the real data distribution $p(x_0)$ into its noised versions  $x_1, \dots,  x_T$ through a Markov chain of length $T$, as: 
\begin{equation}
    q(x_t | x_{t-1}) = \mathcal{N}(x_t; \sqrt{\alpha_t}  x_{t-1}, (1 - \alpha_t)\mathbf{I}),
    \label{eq:noising_process}
\end{equation}
for timesteps $t = 1, \dots, T$, where $\alpha_1, \dots,  \alpha_T$ represent a fixed variance schedule.
However, any step of the noised sample can also be efficiently produced using a closed-form expression directly from the input $x_0$~\cite{ho2020denoising} as follows:
\begin{equation}
    x_{t} = \sqrt{\bar{\alpha}_t}x_{0} + (1 - \bar{\alpha}_t) \epsilon, \quad \epsilon \sim \mathcal{N}(0, \mathbf{I})
    \label{eq:noising_directly}
\end{equation}
with $\bar{\alpha}_t := \prod^{t}_{s=1}{\alpha_s}$, which enables uniform sampling of $t$. 
Through training, the denoising network (i.e., typically a U-Net network~\cite{ronneberger2015unet}), learns to estimate the real data distribution from a noise-filled standard Gaussian distribution. This entails gradually denoising a noisy image $x_T \sim \mathcal{N}(0, \textbf{I})$ to less noisy samples $x_t$ until a denoised data sample $x_0$ is reached. To this end, the denoising network $\epsilon_\theta(x_t, t)$ predicts the noise $\epsilon$ that is added at step $t$ with Equation~\eqref{eq:noising_directly}.


\vspace{0.8mm}\noindent\textbf{Variational Auto-Encoder (VAE).}~To greatly improve efficiency, the noising and denoising processes are carried out in the  latent space of a pretrained Variational Auto-Encoder~(VAE) instead of the pixel space~\cite{rombach2022high_latent_diffusion}. This is achieved by first mapping the input sample $x_0$ to the latent input $z_0$ through the encoder model $\mathcal{E}$ of the VAE. Noising with Equation~\eqref{eq:noising_directly} is then performed to obtain noised samples $z_t$ on which the denoising network $\epsilon_{\theta}$ is trained.  
During inference, synthetic images can then be generated by randomly sampling a noisy sample $z_T$ in the latent space, denoising it with the predictor $\epsilon_{\theta}$, and then mapping the denoised sample $z_0$ back to the pixel space with the VAE decoder $\mathcal{D}$.

\vspace{0.8mm}\noindent\textbf{Text encoder.}
To enable control over the generation process, the denoising network is also conditioned on input text prompts~\cite{ho2020denoising}. The text prompt is first tokenized and mapped to corresponding token embeddings, which are then encoded through a pretrained CLIP text encoder~\cite{radford2021learning_CLIP}. Encoded prompts $c$ are then passed as conditions to the denoising network through the cross-attention mechanism~\cite{chen2021crossvit}.







\subsection{Training objectives of ID-Booth}
Pretrained diffusion models provide unparalleled text-guided synthesis capabilities, owing to training on various datasets of unprecedented scale~\cite{rombach2022high_latent_diffusion}. 
However, their knowledge of very specific concepts and styles remains limited. This is also true for their ability to create images of a desired identity as prompting for a specific non-celebrity identity can be difficult or even impossible.

To facilitate the generation of identity-specific images, we propose to fine-tune a pretrained diffusion model on a small set of input images of a desired identity. Our proposed ID-Booth framework utilizes three separate training objectives, to improve identity consistency while retaining the synthesis capabilities of pretrained models.~This includes the reconstruction loss $\mathcal{L}_{REC}$, the prior preservation loss $\mathcal{L}_{PR}$ and a triplet-identity loss $\mathcal{L}_{TID}$, which are combined as follows to form the overall objective:
\begin{equation}  
    \mathcal{L}_{Total} = \mathcal{L}_{REC} + \lambda_{PR}\mathcal{L}_{PR} + \lambda_{TID} \mathcal{L}_{TID},
\label{eq:id-booth_loss}
\end{equation}
as illustrated in Figure~\ref{fig:ID-Booth_framework}. Here, the balancing weight $\lambda_{PR}$ is set to $1.0$, while $\lambda_{TID}$ is defined as $(1 - \frac{t}{T})^2$ to reduce the influence of identity supervision at higher timesteps, as image blurriness increases. The training objectives are described in detail below.  

To further retain the capabilities of pretrained models, while still enabling fine-tuning on new identities, our ID-Booth framework also relies on the use of the Low-Rank Adaptation method (LoRA)~\cite{hu2022lora}. Thus, instead of fine-tuning the entire diffusion model, all existing weights remain frozen while new low-rank trainable layers are introduced in the denoising network. This allows for better retention of synthesis capabilities, while enabling faster training and more efficient storage of fine-tuned model weights. 



\vspace{0.8mm}\noindent\textbf{Reconstruction loss.}~The first training objective of our ID-Booth framework is aimed at image reconstruction and is based on the reweighted optimization objective conventionally used for training diffusion models~\cite{ho2020denoising}. Since denoising is performed in the latent space of a pretrained VAE, the loss is based on the noise $\epsilon$ that is added to sample $z_0$ at timestep $t$ and the noise that is estimated by the denoising network $\epsilon_\theta$ considering the noisy latent sample $z_t$, the timestep $t$, and the text prompt condition $c$. Formally, this reconstruction loss $\mathcal{L}_{REC}$ can be defined as follows:   
\begin{equation}
    \mathcal{L}_{REC} = \mathbb{E}_{z \sim \mathcal{E}(x), \epsilon \sim \mathcal{N}(0, 1), t, c} \bigl [ \Vert \mathbf{\epsilon} - \epsilon_\theta (z_t, t, c)  \Vert_{2}^{2} \bigr].
    \label{eq:LDM_loss}
\end{equation}

\vspace{0.8mm}\noindent\textbf{Prior preservation loss.}~Fine-tuning a diffusion model on a small set of images with only  the reconstruction objective $\mathcal{L}_{REC}$ often leads to overfitting on input data and the loss of prior knowledge, e.g., the concept of what a person is. To address this, our ID-Booth framework utilizes an additional training objective aimed at the preservation of prior concepts~\cite{ruiz2023dreambooth}. To this end, a set of prior images $x_{pr,0}$ are generated by the initial pretrained model prior to training, with prompts related to the novel concept to be introduced. Following the initial reconstruction objective, these prior samples are used to form the prior preservation loss $\mathcal{L}_{PR}$ following the DreamBooth approach~\cite{ruiz2023dreambooth}: 
\begin{equation}
    \mathcal{L}_{PR} = 
    \mathbb{E}_{z_{pr}, c_{pr}, \epsilon', t'} \bigl [ \epsilon_{pr} -  \epsilon_{\theta}(z_{pr,t'}, t', c_{pr}) \Vert_{2}^{2} \bigr],
    \label{eq:DreamBooth_loss}
\end{equation}
where the $pr$ notation represents factors related to prior images generated with the initial model.

\vspace{0.8mm}\noindent\textbf{Triplet identity loss.}~Despite the suitability of $\mathcal{L}_{REC}$ and $\mathcal{L}_{PR}$ for fine-tuning, both objectives are focused solely on image reconstruction and do not target the consistency of generated identities. This is the case for both consistency with desired input identities and consistency among generated samples.
To address this, we propose to incorporate the identity aspect into the training process through the similarity of identity features extracted from images with a pretrained face recognition model. 
However, to enable the inspection of generated identities during training, suitable face images must be produced at each training step. To this end, we use the predicted noise $\epsilon_\theta(z_t, t, c)$ and the latent noisy sample $z_t$ to estimate the denoised latent $\hat{z}_0$ as: 
\begin{equation} 
    \hat{z}_0 = \frac{z_t - \sqrt{1 - \bar{\alpha}_t} \epsilon_\theta}{\sqrt{\bar{\alpha}_t}}.
    \label{eq:denoise_to_start}
\end{equation}
Afterward, we can decode the estimated denoised latent $\hat{z}_0$ to the estimated input image with $\hat{x}_0 = \mathcal{D}(\hat{z}_0)$.
Next, we extract the facial region with a face detection model for both the estimated and the input training image, denoted as $\hat{x}^{f}_{0}$ and $x^{f}_{0}$ respectively. If the facial region exist, we obtain the identity feature representations of each image with a pretrained face recognition model $\varphi$, otherwise the objective is skipped.

To guide the generative model toward better identity consistency, we propose to form a triplet identity objective. The objective utilizes identity features of the reconstructed sample $\hat{x}_0$ as the anchor,  the input image $x_0$ as a positive example of an identity and prior images $x_{pr,0}$ as a negative example.
Formally, our proposed triplet identity objective $\mathcal{L}_{TID}$ can be defined using cosine similarity $cos$ as follows:
\begin{equation}
\scriptstyle    
    \mathcal{L}_{TID} = max\{cos(\varphi(x^{f}_{0}), \varphi(\hat{x}^{f}_{0})) - cos(\varphi(x^{f}_{pr,0}), \varphi(\hat{x}^{f}_{0})) + m, 0\},
\label{eq:triplet_identity_loss}
\end{equation}
where the notations introduced before apply. In addition, $m$ represents a non-negative margin, i.e., the minimum difference between positive and negative similarities that is required for the loss to be zero.
The proposed triplet-objectives also addresses the risk of overfitting on unintentional characteristics of training samples, e.g., the pose, age, hair or accessories, which might leak into the identity embeddings. This is achieved through negative identity examples, which often share similar characteristics with positive examples.

\section{Experiments and Results~\label{sec:experiments_and_results}}
\noindent\textbf{Dataset preparation.}~To fine-tune ID-Booth, we utilize the Tufts Face Database (TFD)~\cite{TUFTS_panetta2018comprehensive}, which contains images captured in a constrained laboratory setting with the consent of subjects. In total, the dataset includes over $10,000$ images of $113$ human subjects captured  across various light spectra. We focus on images captured with four visible field cameras under constant diffused light in a semi-circle around the subjects. During preprocessing, we remove heavily blurred images and extreme side-profile images lacking key facial features (e.g., two eyes), then crop them to focus on the face region, resulting in $2299$ images of $107$ subjects. Next, we use eye landmarks, detected with the Multi-Task Cascaded Convolutional Neural Network~(MTCNN)~\cite{zhang2016MTCNN} to align the faces through an affine transform, and then resize the images to $512\times512$. For evaluation we rely on the Flickr Faces High-Quality~(FFHQ)~\cite{stylegan_1_karras2019style} dataset of $70,000$ diverse in-the-wild unlabeled face images, which we also resize to $512\times512$.

\vspace{0.8mm}\noindent\textbf{Implementation details.} 
We evaluate the suitability of our framework on the state-of-the-art  diffusion model Stable Diffusion 2.1 (SD-2.1)~\cite{rombach2022high_latent_diffusion}, which is capable of generating high-quality and diverse $512\times512$  images through $1000$ denoising timesteps, specified by the discrete denoising scheduler with $\beta_{start}=8.5 \times 10^{-4}$ and $\beta_{end}=0.012$~\cite{ho2020denoising}.  
We fine-tune the SD-2.1 model on images of each identity in the Tufts Face Database (TFD)~\cite{TUFTS_panetta2018comprehensive}. To this end, we utilize the training objectives specified by either DreamBooth~\cite{ruiz2023dreambooth}, focused primarily on image reconstruction, PortraitBooth~\cite{ruiz2023dreambooth}, which includes a simple two-point identity objective, or by our proposed ID-Booth framework, that balances identity consistency and image diversity. The identity objectives utilize features extracted with a pretrained ArcFace-based ResNet-100 recognition model~\cite{deng2019arcface} from face regions detected with MTCNN~\cite{zhang2016MTCNN}.
The detection of faces also acts as the decision factor for when identity-based objectives are applied.
To minimize the effect on the synthesis capabilities of the pretrained model, we utilize the Low-Rank Adaptation (LoRA)~\cite{hu2022lora} method, which freezes the diffusion model but introduces new trainable layers instead. Specifically, we add two linear layers of rank $4$ to each cross-attention block, initialized with a Gaussian distribution. We also generate $200$ images with the initial SD-2.1 model and the prompt \verb|photo of a person|, which are used for preservation of prior concepts through $\mathcal{L}_{PR}$~\cite{ruiz2023dreambooth}.
We then perform fine-tuning with images of a desired identity and the prompt \verb|photo of [ID] person|, where \verb|[ID]| represents a rare text token that will be tied to the new identity, in our case \verb|sks|~\cite{ruiz2023dreambooth}. We utilize an initial learning rate of $10^{-4}$  and the AdamW optimizer~\cite{loshchilov2019decoupled} with $\beta_1 = 0.9$, $\beta_2 = 0.999$, $\epsilon = 10^{-8}$ and a weight decay of $0.01$, along with the half-precision floating point format to lower VRAM usage.  
Fine-tuning is stopped after $32$ epochs (i.e., $6400$ steps), based on our initial observations and existing works~\cite{ruiz2023dreambooth,peng2024portraitbooth}.


\vspace{0.8mm}\noindent\textbf{Data generation.}~\label{sec:prompt_definition}
Each fine-tuned SD model is used to generate two synthetic datasets, one with $21$ images per identity, as is the case in TFD~\cite{TUFTS_panetta2018comprehensive}, and one with $100$ images per identity to investigate the scalability of our approach. Data generation is performed with a guidance scale of $5.0$ and $30$ inference denoising steps with the same discrete denoising scheduler as during training. The goal is to generate diverse synthetic images of desired identities under various scenarios. To produce images that resemble real-world in-the-wild datasets~\cite{stylegan_1_karras2019style}, we utilize a prompt that defines a face image of a specific identity as well as the environment the image is taken in: 
\begin{lstlisting}[breaklines, basicstyle=\footnotesize]
face [P] photo of [G] [ID] person, [B] background
\end{lstlisting}
Here \verb|[ID]| represents the identity token, while \verb|[G]| defines the gender of the person, i.e., \verb|female| or \verb|male|. To generate diverse images we also select the environment through \verb|[B]|, which we randomly sample from the following list: 
\begin{lstlisting}[breaklines, basicstyle=\footnotesize, breakindent=0pt]
forest, city street,  bus, office, factory, beach, laboratory, construction site, hospital, night club
\end{lstlisting}
To also produce a variety of poses we randomly select whether the image is a \verb|portrait| or a \verb|side-portrait|, represented by \verb|[P]|. 
In addition, we rely on the following negative prompt to obtain more realistic images:
\begin{lstlisting}[breaklines, basicstyle=\footnotesize, breakindent=0pt]
cartoon, render, illustration, painting, drawing, black and white, bad body proportions, landscape 
\end{lstlisting}
An ablation study of the main prompt components is available in the supplementary material. 


\begin{table*}[t!] 
\caption{\textbf{Quantitative evaluation of quality, fidelity and diversity of synthetic images.} Quality is assessed with Fréchet Distance~\cite{heusel2017gans_FID} and Kernel Distance~\cite{binkowski2018demystifying_kernel_distance}, while fidelity and diversity are measured through Density and Coverage~\cite{naeem2020reliable}. Results are computed by comparing distributions of features extracted with DINOv2-ViT-L/14~\cite{oquab_2024_dinov2} from synthetic images and real-world images of FFHQ~\cite{stylegan_1_karras2019style}, considering either entire images or only the face region. Vendi score~\cite{friedman_2024_vendi} is used to evaluate intra-identity diversity, while CR-FIQA~\cite{boutros2023cr_fiqa} measures face image quality of each sample, both without a reference dataset. 
\vspace{-2mm}
}
\centering
\resizebox{\linewidth}{!}{%
\begin{tabular}{llccccccccccc}
\toprule
 &  & \multicolumn{2}{c}{\bf{Fréchet Distance} $\downarrow$} &   \multicolumn{2}{c}{\bf{Kernel Distance} $\downarrow$}  &  \multicolumn{2}{c}{\bf{Density} $\uparrow$}  & \multicolumn{2}{c}{\bf{Coverage} $\uparrow$} & \multicolumn{2}{c}{\bf{Vendi score per ID $\uparrow$}}  & \bf{CR-FIQA}   \\ 
\bf{Data from} & \bf{Method}  &  Entire &  Face  &  Entire &  Face &  Entire &  Face & Entire & Face & Entire & Face & Face region  \\ 
\midrule  
TFD~\cite{TUFTS_panetta2018comprehensive} & Real data & $2035.615$ & $1679.317$ &  $7.056$ & $5.779$ & $0.195$ & $0.623$ & $0.043$ & $0.120$ & $2.536$ & $3.132$  & $2.131 \pm 0.094$ \\
FFHQ~\cite{stylegan_1_karras2019style} & Real data & $38.703$ & $33.799$ & $0.001$ & $0.001$ & $1.028$ & $1.007$ & $0.972$ & $0.970$ & $-$ & $-$ & $2.090 \pm 0.134$ \\ 

\midrule

\multirow{5}{*}{{SD-2.1}} & No fine-tuning & $1123.226$ & $1075.064$ & $2.201$ & $2.881$ & $0.422$ & $0.413$ & $0.137$ & $0.204$ & $-$ & $-$ &  $2.080 \pm 0.238$ \\
\cmidrule{2-13}
& DreamBooth~\cite{ruiz2023dreambooth} & $1374.696$ & $1371.129$ & $4.134$ & $4.484$ & \bm{$0.698$} & \bm{$0.575$} & $0.128$ & $0.131$ & $7.264$ & $6.705$ & ${2.187} \pm 0.134$\\ 
& PortraitBooth~\cite{peng2024portraitbooth} & \underline{$1182.511$} & \underline{$1202.684$} & \underline{$3.000$} & \underline{$3.568$} & \underline{$0.575$} & \underline{$0.510$} & \underline{$0.149$} & \underline{$0.154$} & \underline{$12.192$} & \underline{$9.614$} & {$2.149  \pm 0.173$}\\
& ID-Booth (ours) & \bm{$1144.651$} & \bm{$1159.537$} & \bm{$2.778$} & \bm{$3.346$} & $0.536$ & $0.502$ & \bm{$0.157$} & \bm{$0.166$} & \bm{$13.510$} & \bm{$10.430$} & $2.143 \pm {0.181}$\\ 

\bottomrule

\multicolumn{13}{l}{($\downarrow$ / $\uparrow$) -- Lower / Higher is better; (\textbf{Bold}) -- Best result; (\underline{Underline}) -- Second best result} \\
\end{tabular}
} 
\vspace{-2mm}
\label{tab:quality_fidelity_diversity}
\end{table*}

\begin{table}[t!] 
\caption{\textbf{Evaluation of diversity through pose estimation.} Reported are the mean and standard deviation of standard deviation values of pitch, yaw and roll measured across samples of each identity with the 6DRepNet~\cite{hempel2024toward} head pose estimator.
\vspace{-2mm}
}
\centering
\resizebox{\linewidth}{!}{%
\begin{tabular}{llccc}
\toprule
 &  &   \multicolumn{3}{c}{\bf{Pose estimation }}    \\ 
       \bf{Data from} & \bf{Method} &  \bf{Pitch ($\sigma$ per ID)}  & \bf{Yaw ($\sigma$ per ID)}  & \bf{Roll ($\sigma$ per ID)} \\ 
    \midrule  
TFD~\cite{TUFTS_panetta2018comprehensive} & Real data & $2.015 \pm 0.718$  & $26.297 \pm 4.609$  & $2.285 \pm 1.114$ \\
\midrule
\multirow{5}{*}{{SD-2.1}} & No fine-tuning & $7.486 \pm 1.487$  & $24.909 \pm 2.596$  & $4.835 \pm 1.656$ \\ 
\cmidrule{2-5}
& DreamBooth~\cite{ruiz2023dreambooth} & $4.681 \pm 1.232$  & $26.118 \pm 5.261$  & $3.248 \pm 1.449$ \\
& PortraitBooth~\cite{peng2024portraitbooth}  & $6.249 \pm 2.423$  & $33.159 \pm 6.735$  & $5.241 \pm 2.563$ \\
& ID-Booth (ours) & \bm{$6.641 \pm 2.662$}  & \bm{$33.527 \pm 7.569$}  & \bm{$5.637 \pm 2.920$} \\ 

\bottomrule

\multicolumn{5}{l}{($\downarrow$ / $\uparrow$) -- Lower / Higher is better; (\textbf{Bold}) -- Best result; (\underline{Underline}) -- Second best result} \\
\end{tabular}
} 
\vspace{-2mm}
\label{tab:pose_estimation}
\end{table}

\vspace{0.8mm}\noindent\textbf{Evaluation methodology.} We evaluate the suitability of the proposed ID-Booth framework by comparing its synthesis capabilities to those of DreamBooth~\cite{ruiz2023dreambooth} and a version of it extended with the PortraitBooth~\cite{peng2024portraitbooth} identity objective. Other diffusion-based frameworks that produce identity-specific images, e.g., Arc2Face~\cite{papantoniou2024arc2face} and InstantID~\cite{wang2024instantid}, are not considered as they are trained on large-scale web-scraped face recognition datasets, without the consent of subjects. Meanwhile, our experiments entail fine-tuning on a limited amount of images from TFD~\cite{TUFTS_panetta2018comprehensive} gathered with suitable consent. To evaluate the produced images, we compare them to the 
diverse real-world images of FFHQ~\cite{stylegan_1_karras2019style}. Here, we consider either entire images or only the face regions of a resolution $112 \times 112$, aligned and cropped based on face landmarks detected by MTCNN~\cite{zhang2016MTCNN}. 
The quality of images is then determined with Fréchet Distance~\cite{heusel2017gans_FID} and Kernel Distance~\cite{binkowski2018demystifying_kernel_distance}, measured on features extracted with the pretrained DINOv2-ViT-L/14 model~\cite{oquab_2024_dinov2} rather than the typical Inception-v3~\cite{szegedy2016rethinking_inception} model, which has been shown to be unsuitable, due to poor correlation with human evaluators~\cite{stein2023exposing_dgm_eval} and the limitations of the ImageNet dataset~\cite{deng2009imagenet}. 
We also evaluate the fidelity and diversity of images separately, with the use of Density and Coverage~\cite{naeem2020reliable}, measured on features of DINOv2-ViT-L/14~\cite{oquab_2024_dinov2}. To compute these scores we utilize the generated datasets with $100$ samples per identity and compare them to $10.000$ samples from FFHQ~\cite{stylegan_1_karras2019style}.
In addition, we analyze the intra-identity diversity of samples via the extracted features with the Vendi score~\cite{friedman_2024_vendi}, which differently from previous measures does not require a reference dataset.  
Similarly, we rely on the Certainty Ratio Face Image Quality Assessment (CR-FIQA)~\cite{boutros2023cr_fiqa} to evaluate the quality of each face image through relative classifiability with a pretrained ResNet-101~\cite{he2016deep_resnet}.
We also analyze intra-identity diversity by evaluating the pitch, yaw and roll of faces in the images  with the 6DRepNet~\cite{hempel2024toward} head pose estimator. 

\vspace{0.8mm}\noindent\textbf{Recognition experiment details.}~\label{sec:details_recognition_experiments}
As part of our experiments, we also investigate intra-identity consistency and inter-identity separability based on genuine and imposter distributions. These are formed using the cosine similarity of identity features of synthetic samples and either samples of the corresponding identity (genuine pair) or a different identity (imposter pair), from either TFD~\cite{TUFTS_panetta2018comprehensive} or the synthetic dataset. The identity features are extracted with a Resnet-101~\cite{he2016deep_resnet} recognition model trained with the ArcFace loss~\cite{deng2019arcface} on the MSV1MV3 dataset~\cite{guo2016msceleb} from face regions of the images detected with MTCNN~\cite{zhang2016MTCNN}. To allow for a fair comparison with samples of TFD~\cite{TUFTS_panetta2018comprehensive}, we form the distributions with synthetic datasets that also consist of $21$ samples per identity.
For each dataset combination, we form all possible genuine pairs along with an equal amount of randomly sampled imposter pairs. 
We report the mean and standard deviation of distributions along with established metrics, including Equal Error Rate (EER), False Match Rate at a False Non-Match Rate of $1.0\%$ (FMR100) or $0.01\%$ (FMR1000), False Non-Match Rate at a False Match Rate of $1.0\%$ (FNMR100) or $0.01\%$ (FNMR1000), and the Fisher Discriminant Ratio~(FDR)~\cite{iso_iec_biometric_standards}.
Lastly, we use the produced data to augment the TFD~\cite{TUFTS_panetta2018comprehensive} dataset, which is then used to train a ResNet-50~\cite{he2016deep_resnet} recognition model with the AdaFace loss~\cite{kim2022adaface}. For training we utilize a batch size of $128$ and the Stochastic Gradient Descent (SGD) optimizer with $0.9$ momentum, a weight decay of $5 \times 10^{-4}$, and a dropout ratio of $0.4$. The learning rate is initially set to $0.1$ and is lowered by a factor of $10$ after the 22nd, the 30th, and the 35th epoch. Training is stopped once no improvement in $5$ epochs is observed  on the LFW~\cite{huang2008labeled_LFW} benchmark. The performance of the trained model is then evaluated on five state-of-the-art verification benchmarks, including Labeled Faces in the Wild (LFW)~\cite{huang2008labeled_LFW}, its Cross-Age and Cross-Pose subsets CA-LFW~\cite{zheng2017cross_CALFW} and CP-LFW~\cite{zheng2018cross_CPLFW}, Celebrities in Frontal-Profile in the Wild (CFP-FP)~\cite{sengupta2016frontal_CFPFP} and AgeDB-30~\cite{moschoglou2017agedb}.

\noindent\textbf{Experimental hardware.}
The experiments were conducted on a cluster of $4$ Nvidia A100 SXM4 40GB GPUs and a Desktop PC with an Nvidia RTX $4090$ GPU.

\begin{figure}[tb!]
    \centering
    \includegraphics[width=\linewidth]{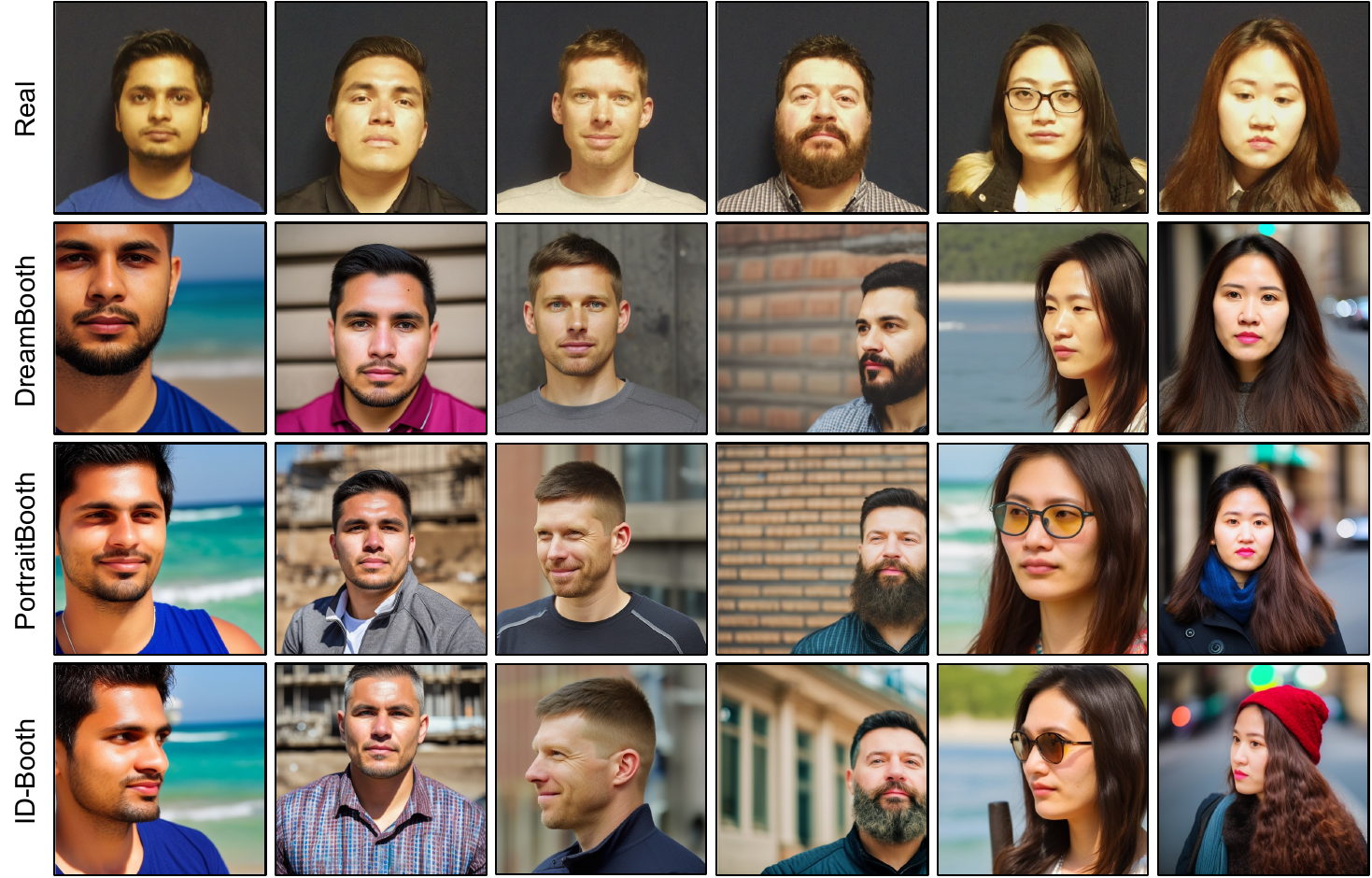}
    \vspace{-6mm}\caption{\textbf{Comparison of generated image samples.} ID-Booth facilitates better identity consistency than DreamBooth~\cite{ruiz2023dreambooth} and better image diversity than when utilizing the PortraitBooth~\cite{peng2024portraitbooth} identity objective, which can limit the variety of facial features and poses.
    } 
    \label{fig:diversity_and_consistency_samples}
    \vspace{-6mm}
\end{figure}

\subsection{Evaluation of generated images}~\label{sec:image_evaluation}

\vspace{-4mm}\noindent\textbf{Image quality.}~We begin our evaluation by assessing the overall quality of images produced by either the proposed ID-Booth framework or its two competitors, the base DreamBooth~\cite{ruiz2023dreambooth} and a version of it extended with the  PortraitBooth~\cite{peng2024portraitbooth} identity objective (denoted as PortraitBooth for brevity).
To this end, we utilize the Fréchet Distance~\cite{heusel2017gans_FID} and Kernel Distance~\cite{binkowski2018demystifying_kernel_distance} computed between features extracted with DINOv2-ViT-L/14~\cite{oquab_2024_dinov2} from synthetic images and features extracted from real-world images of the FFHQ~\cite{stylegan_1_karras2019style} dataset. From results reported in Table~\ref{tab:quality_fidelity_diversity} and samples in Figures~\ref{fig:IDBooth_diverse_samples} and \ref{fig:diversity_and_consistency_samples}, we can discern that with the SD-2.1 model and our defined prompts we can generate images that better match the quality of in-the-wild FFHQ~\cite{stylegan_1_karras2019style} images than the real-world images of TFD~\cite{TUFTS_panetta2018comprehensive}, which were gathered in a constrained environment. Comparing results of the different fine-tuning methods, we see that our ID-Booth framework achieves the best quality results, scoring closest to the non fine-tuned model, while enabling identity-specific generation. This is the case both when evaluating entire images or only the cropped face regions.
Interestingly, both identity-based training objectives of either PortraitBooth~\cite{peng2024portraitbooth} or ID-Booth improve the image quality of the base DreamBooth~\cite{ruiz2023dreambooth}. 

In addition, we evaluate the quality of each face region with CR-FIQA~\cite{boutros2023cr_fiqa}. Here, however, high face quality is not necessarily as desired as having a mix of high and low quality images, which can lead to better performing recognition models. This difference can also be observed on real-world datasets, where in-the-wild images of FFHQ~\cite{stylegan_1_karras2019style} achieve a lower mean but higher standard deviation than constrained images of TFD~\cite{TUFTS_panetta2018comprehensive}. 
Similarly, compared to DreamBooth~\cite{ruiz2023dreambooth} and the PortraitBooth-based~\cite{peng2024portraitbooth} version, our ID-Booth framework achieves a higher standard deviation but lower mean of CR-FIQA~\cite{boutros2023cr_fiqa} scores, that are closer to those of diverse samples of the non fine-tuned models.

\begin{figure}[tb!]
    \centering
    \includegraphics[width=\linewidth]{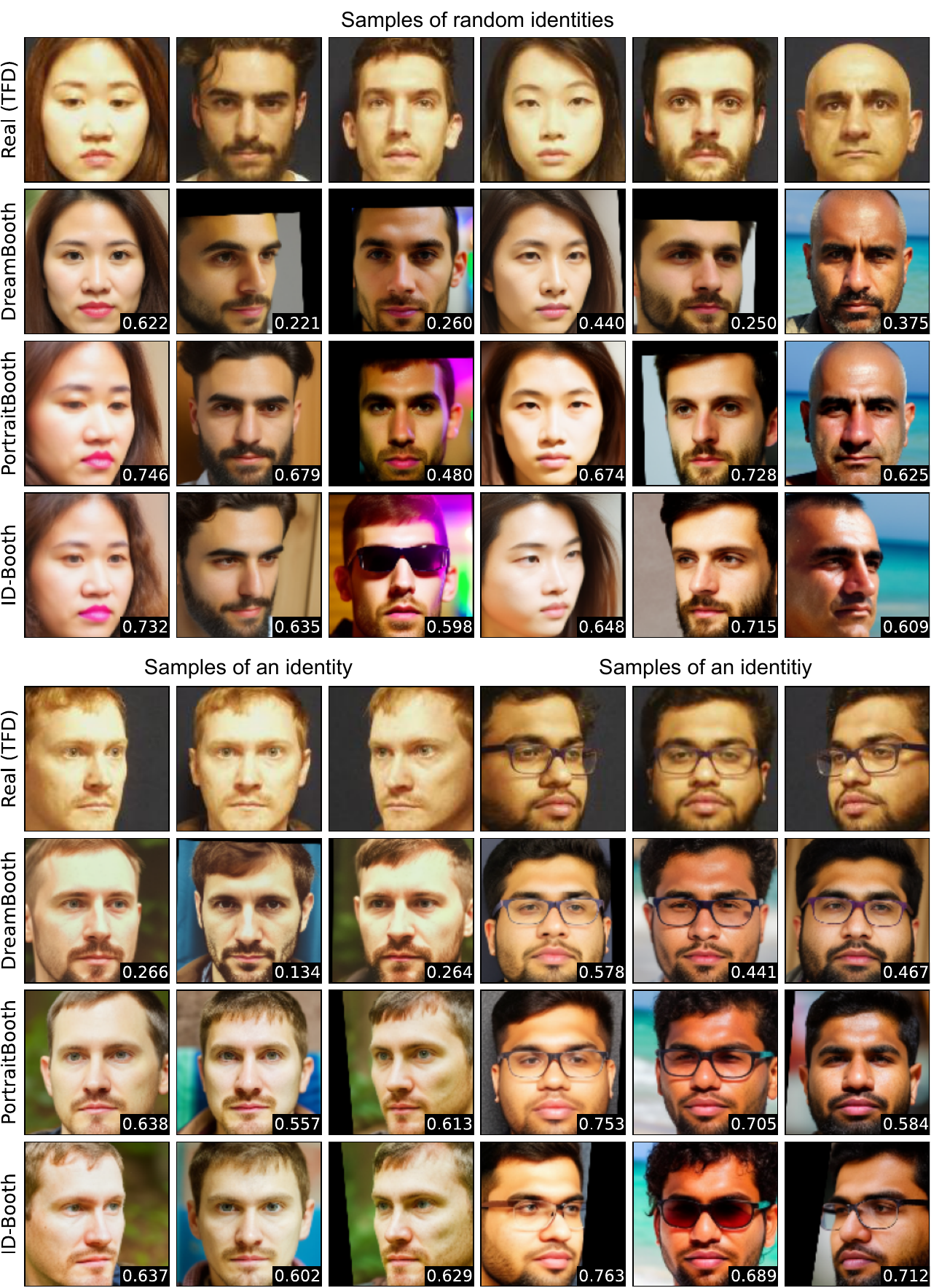}
    \vspace{-6mm}
    \caption{\textbf{Comparison of identity consistency.} ID-Booth achieves better identity consistency than DreamBooth~\cite{ruiz2023dreambooth}, while retaining more diverse synthesis capabilities and ensuring better intra-identity diversity than PortraitBooth~\cite{peng2024portraitbooth}. Reported is the cosine similarity of synthetic and real identity features extracted with the pretrained ArcFace recognition model~\cite{deng2019arcface}.
    } 
    \label{fig:identity_consistency_samples}
    \vspace{-6mm}
\end{figure}

\vspace{0.8mm}\noindent\textbf{Image fidelity and diversity.}~Next, we analyze the produced images in terms of fidelity, i.e., the degree to which they resemble real samples, and diversity, i.e., how well they cover the variability of real samples~\cite{sajjadi2018assessing}. 
To this end, we rely on Density and Coverage~\cite{naeem2020reliable}, respectively, reported in Table~\ref{tab:quality_fidelity_diversity}, in addition to qualitative samples in Figures~\ref{fig:diversity_and_consistency_samples} and~\ref{fig:identity_consistency_samples}.
Comparing different datasets with FFHQ~\cite{stylegan_1_karras2019style}, we can observe that images of TFD~\cite{TUFTS_panetta2018comprehensive} lack the fidelity and diversity expected of in-the-wild images. Synthetic samples produced by the non-finetuned SD-2.1 model with diverse prompts offer a notable improvement in these areas.
In comparison, fine-tuning approaches achieve a higher fidelity of entire images and face regions, but crucially result in lower diversity of face regions.   
Among the approaches, DreamBooth~\cite{ruiz2023dreambooth} scores the highest in terms of density (i.e., fidelity), while ID-Booth achieves the highest coverage (i.e., diversity) on both entire images and detected face regions. This can also be observed in Figures~\ref{fig:diversity_and_consistency_samples} and~\ref{fig:identity_consistency_samples}, where ID-Booth offers more consistent identities, while enabling a larger variety of poses, ages, accessories and other facial features than DreamBooth~\cite{ruiz2023dreambooth} or the PortraitBooth-based~\cite{peng2024portraitbooth} version of it.

\vspace{0.8mm}\noindent\textbf{Intra-identity diversity.} To further investigate the produced images, we also analyze the intra-diversity of samples with the per-class Vendi score~\cite{friedman_2024_vendi}. As reported in Table~\ref{tab:quality_fidelity_diversity}, synthetic images generated by all fine-tuning methods offer more intra-identity diversity than the constrained samples of TFD~\cite{TUFTS_panetta2018comprehensive}. Our ID-Booth achieves the largest intra-identity diversity among the fine-tuning approaches, both of entire images and only face regions, while ensuring better identity consistency. This can be seen in Figure~\ref{fig:identity_consistency_samples}, where ID-Booth samples of the same identity contain a larger variety of poses and face accessories. 
To obtain deeper insight, we also analyze the pitch, yaw and roll of faces with the 6DRepNet~\cite{hempel2024toward} head pose estimator. In Table~\ref{tab:pose_estimation} we report the mean and standard deviation of standard deviation values obtained from pose distributions of each identity. Results reveal that ID-Booth generates samples with the largest variety of poses per identity, especially in terms of pitch and roll, which represents an important aspect of overall intra-identity diversity. In comparison, DreamBooth~\cite{ruiz2023dreambooth} and the PortraitBooth-based~\cite{peng2024portraitbooth} version often default to more front-facing poses, as seen in Figures~\ref{fig:diversity_and_consistency_samples} and~\ref{fig:identity_consistency_samples}.


\subsection{Recognition-based experiments}\label{sec:recognition_experiments}

\noindent\textbf{Identity consistency and separability.}~To determine the suitability of generated images for augmenting recognition datasets we must also examine the consistency and separability of identities in the images. 
To this end, we form genuine and imposter distributions either only among synthetic identities or between synthetic and real-world identities, based on the similarity of features extracted with the pretrained ArcFace-based recognition model~\cite{deng2019arcface}. 
From verification scores in Table~\ref{tab:pyeer_benchmarks_vs_real_and_vs_synthetic} that describe these distributions, we can discern that the use of identity-based training objectives from either ID-Booth or PortraitBooth~\cite{peng2024portraitbooth} results in better identity consistency and separability than the base DreamBooth~\cite{ruiz2023dreambooth}. This is the case both either among synthetic or between synthetic and real identities across all verification measures. The only exception is the FNMR1000 score in the latter scenario, which can likely be attributed to a handful of outliers, especially when considering the lower FNMR100 scores and improved FDR values.  Compared to the PortraitBooth-based~\cite{peng2024portraitbooth} approach, ID-Booth achieves lower FNMR scores among synthetic identities along with lower FMR and FNMR scores between synthetic and real identities, indicating fewer outliers. 
In combination with a higher FDR score in the second scenario, this signifies better intra-identity consistency and inter-identity separability. 
ID-Booth samples in Figure~\ref{fig:identity_consistency_samples} further support these observations with better consistency between generated and real identities or among different samples of the same identity. 
Overall, these results highlight crucial characteristics of ID-Booth, demonstrating its suitability for augmenting existing datasets in a privacy-preserving manner by producing identity-consistent in-the-wild images of real-world identities from the training dataset collected with subject consent.


\begin{table*}[tb!] 
\caption{\textbf{Evaluation of consistency and separability between synthetic and real-world identities.}
Reported are verification measures of genuine and imposter distributions among synthetic or between synthetic and real images, constructed based on the cosine similarity of identity features obtained with a pretrained ArcFace-based recognition model~\cite{deng2019arcface}.
\vspace{-2mm}
}
\centering
\resizebox{0.9\linewidth}{!}{%
\begin{tabular}{llcccccc}
\toprule
    \bf{Data setting}  & \bf{Method} & \bf{EER $\downarrow$} & \bf{FMR100 / 1000 $\downarrow$} & \bf{FNMR100 / 1000 $\downarrow$}  & \bf{Imposter $\mu \pm \sigma$ $\downarrow$} &  \bf{Genuine $\mu \pm \sigma$ $\uparrow$} & \bf{FDR $\uparrow$} \\ 
    \midrule
among TFD~\cite{TUFTS_panetta2018comprehensive} & Real data  & $0.002$ & $0.002$ / $0.002$ &  $0.001$ / $0.003$  & $0.021 \pm 0.0725$ & $0.871 \pm 0.070$ & $70.969$ \\
\midrule 

\multirow{3}{*}{among SD-2.1} & DreamBooth~\cite{ruiz2023dreambooth} & $0.055$ & $0.153$ / $0.337$ & $0.297$ / \underline{$0.919$} & $0.103 \pm 0.093$ & \bm{$0.499 \pm 0.141$} & $5.509$ \\
 & PortraitBooth~\cite{peng2024portraitbooth} & \underline{$0.043$} & \underline{$0.096$} / \underline{$0.230$} & \underline{$0.269$} / $0.967$ & \underline{$0.065 \pm 0.084$} & \underline{$0.495 \pm 0.151$} & \bm{$6.210$} \\
 & ID-Booth (ours) & \bm{$0.042$} & \bm{$0.095$} / \bm{$0.217$} & \bm{$0.249$} / \bm{$0.896$} & \bm{$0.059 \pm 0.082$} & $0.486 \pm 0.153$ & \underline{$6.073$} \\
\midrule 
\multirow{3}{*}{SD-2.1 vs. TFD} & DreamBooth~\cite{ruiz2023dreambooth} & $0.046$ & $0.087$ / $0.184$ & $0.286$ / \bm{$0.684$} & $0.019 \pm 0.072$ & $0.406 \pm 0.155$ & $5.132$ \\
 & PortraitBooth~\cite{peng2024portraitbooth} & \underline{$0.028$} & \underline{$0.048$} / \underline{$0.112$} & \underline{$0.133$} / $0.848$ & \underline{$0.017 \pm 0.073$} & \underline{$0.465 \pm 0.153$} & \underline{$6.987$} \\
 & ID-Booth (ours) & \bm{$0.027$} & \bm{$0.044$} / \bm{$0.091$} & \bm{$0.110$} / \underline{$0.838$} & \bm{$0.017 \pm 0.072$} & \bm{$0.465 \pm 0.148$} & \bm{$7.402$} \\
 
\bottomrule  
\multicolumn{6}{l}{($\downarrow$ / $\uparrow$) -- Lower / Higher is better; (\textbf{Bold}) -- Best result; (\underline{Underline}) -- Second best result}
\end{tabular}
} 
\vspace{-2mm}

\label{tab:pyeer_benchmarks_vs_real_and_vs_synthetic}
\end{table*}

\begin{table*}[t!] 
\caption{\textbf{Verification performance of recognition models trained on real and synthetic data.} Reported is the accuracy of a ResNet-50 model trained with AdaFace loss~\cite{kim2022adaface} across $5$ real-world verification benchmarks. LFW~\cite{huang2008labeled_LFW} is used for validation. 
\vspace{-2mm}
}
\centering
\resizebox{0.9\linewidth}{!}{%
\begin{tabular}{llcccccc}
\toprule
\multicolumn{2}{c}{\bf{Training setting}} & \multicolumn{6}{c}{\bf{Verification accuracy on benchmarks $\uparrow$} } \\
\cmidrule(lr){1-2} \cmidrule(lr){3-8} 
\bf{Dataset (samples per ID)} & \bf{Method}& \bf{LFW} & \bf{AgeDB-30} & \bf{CA-LFW} & \bf{CFP-FP}  & \bf{CP-LFW}  & \bf{Average accuracy} \\ 
\cmidrule(lr){1-2} \cmidrule(lr){3-8} 
TFD (21) &  Real data &  $0.688$ & $0.500$ & $0.557$ & $0.593$ & $0.540$ & $0.575 \pm 0.063$  \\ 


\cmidrule(lr){1-2} \cmidrule(lr){3-8} 
\multirow{3}{*}{TFD (21) + SD-2.1 (21)} & DreamBooth~\cite{ruiz2023dreambooth} & \underline{$0.755$} & $0.531$ & $0.602$ & $0.603$ & \underline{$0.576$}& $0.613 \pm 0.075$   \\ 
& PortraitBooth~\cite{peng2024portraitbooth} & $0.753$ & \underline{$0.561$} & \bm{$0.615$} & \underline{$0.618$} & $0.576$ & \underline{$0.624 \pm 0.068$}   \\ 
& ID-Booth (ours) & \bm{$0.765$} & \bm{$0.595$} & \underline{$0.612$} & \bm{$0.621$} & \bm{$0.587$} & \bm{$0.636 \pm 0.066$}   \\ 


\cmidrule(lr){1-2} \cmidrule(lr){3-8} 
\multirow{3}{*}{TFD (21) +  SD-2.1 (100)} & DreamBooth~\cite{ruiz2023dreambooth} & $0.766$ & $0.561$ & $0.606$ & \bm{$0.646$} & $0.592$& $0.634 \pm 0.071$   \\ 
& PortraitBooth~\cite{peng2024portraitbooth} & \underline{$0.787$} & \underline{$0.565$} & \underline{$0.627$} & {$0.631$} & \underline{$0.605$} & \underline{$0.643 \pm 0.076$}   \\ 
& ID-Booth (ours) & \bm{$0.790$} & \bm{$0.609$} & \bm{$0.635$} & \underline{$0.632$} & \bm{$0.606$} & \bm{$0.654 \pm 0.069$}   \\

\bottomrule 
\multicolumn{8}{l}{($\uparrow$) -- Higher is better; (\textbf{Bold}) -- Best result; (\underline{Underline}) -- Second best result} 
\end{tabular}
} 
\vspace{-4mm}
\label{tab:FR_benchmarks}
\end{table*}


\vspace{0.8mm}\noindent\textbf{Training face recognition models.}~Lastly, we also explore the real-world utility of the generated data for training deep face recognition models. 
As part of our experiments, we utilize the produced synthetic datasets to augment the small-scale real-world Tufts Face Database~(TFD)~\cite{TUFTS_panetta2018comprehensive} with in-the-wild synthetic images of its identities, whose real-world images were gathered with consent in a constrained laboratory setting.
The augmented datasets are then used to train a ResNet-50~\cite{he2016deep_resnet} recognition model with the AdaFace loss~\cite{kim2022adaface}, following the procedure described in Section~\ref{sec:details_recognition_experiments}.
The suitability of the synthetic datasets obtained with different fine-tuning methods is then evaluated  based on the performance of the trained recognition model on five state-of-the-art face verification benchmarks. 

We first explore augmentation with synthetic datasets consisting of $21$ samples per identity, denoted as SD-2.1~(21), which matches the scale of TFD~\cite{TUFTS_panetta2018comprehensive}. Results reported in Table~\ref{tab:FR_benchmarks} reveal that this form of augmentation enables the training of drastically better performing recognition models compared to training on only real-world samples of TFD~\cite{TUFTS_panetta2018comprehensive}. Among the fine-tuning approaches, our ID-Booth framework achieves the highest overall accuracy, with improvements being especially evident on the AgeDB-30~\cite{moschoglou2017agedb} and CP-LFW~\cite{zheng2018cross_CPLFW} benchmarks. In total, using ID-Booth for augmentation facilitates a $6.1\%$ average accuracy increase over the recognition model trained on the non-augmented dataset.  
In the second set of experiments, we investigate the augmentation capabilities of larger synthetic datasets consisting of $100$ samples per identity, denoted as SD-2.1~(100). From results in Table~\ref{tab:FR_benchmarks} we can discern notable accuracy improvements across with all fine-tuning approaches, showcasing their scalability. Similar to previous results, our ID-Booth framework achieves higher overall verification accuracy than DreamBooth~\cite{ruiz2023dreambooth} or the PortraitBooth-based~\cite{peng2024portraitbooth} version, with a total average augmentation improvement of $7.9\%$. As before, the most noticeable improvement is observed on cross-age benchmarks (i.e., AgeDB-30~\cite{moschoglou2017agedb} and CA-LFW~\cite{zheng2017cross_CALFW}).
Overall, the accuracy improvements obtained with our ID-Booth framework can be attributed to the novel triplet identity objective, which enables better identity consistency and higher intra-identity diversity, especially in terms of age and pose, than either DreamBooth~\cite{ruiz2023dreambooth} or the PortraitBooth-based~\cite{peng2024portraitbooth} version. 
\section{Conclusion}
In this paper, we presented ID-Booth, a new diffusion-based fine-tuning framework for generating high-quality identity-consistent face images, suitable for augmenting existing small-scale recognition datasets in a privacy-preserving manner. 
To this end, ID-Booth relies on a novel triplet identity training objective that improves both intra-identity consistency and inter-identity separability, while retaining image diversity of pretrained state-of-the-art diffusion models. 
Throughout the experiments, we showcase that training deep recognition models on datasets augmented with synthetic samples of ID-Booth results in better performance across five verification benchmarks than when performing augmentation with synthetic samples of existing approaches or training only on real-world data. With regard to future work, we aim to investigate the applicability of identity-based objectives in the training of conditioning approaches and plan to explore the creation of larger-scale datasets.

\section*{ETHICAL IMPACT STATEMENT}


This research primarily focuses on investigating the utility of synthetic data for face recognition tasks, which has potential benefits for privacy-preserving biometric systems~\cite{boutros2023synthetic,meden2021privacy}. The proposed ID-Booth framework is designed to generate high-quality synthetic facial images that ensure identity consistency while preserving image diversity, making them suitable for dataset augmentation. 
The generated data can help mitigate privacy concerns associated with real face datasets by reducing the need for large-scale data collection, which often includes web-scraping without the proper consent of subjects in the images~\cite{jasserand2018massive}. Furthermore, this work limits the need for the distribution and storage of personally identifiable facial images. 

However, the use of generative models for facial image synthesis also carries ethical considerations. The potential misuse of identity-consistent synthetic faces includes unauthorized impersonation, deepfake-related fraud, and misuse in surveillance applications. Furthermore, biases inherent in the training data of latent diffusion models~\cite{rombach2022high_latent_diffusion} could be reflected in the generated images, potentially leading to demographic disparities. 

To mitigate these risks, our work follows strict ethical guidelines and established protocols from prior works on synthetic face generation~\cite{boutros2023synthetic}. The training dataset used in our research consists of images from the Tufts Face Database~\cite{TUFTS_panetta2018comprehensive}, which were captured with the consent of subjects in a constrained laboratory setting. Thus, we do not train on or generate images of real individuals that have not given their consent. Other datasets~\cite{stylegan_1_karras2019style} and benchmarks~\cite{huang2008labeled_LFW,moschoglou2017agedb,sengupta2016frontal_CFPFP,zheng2018cross_CPLFW,zheng2017cross_CALFW} used solely for evaluation are also publicly available, thus ensuring compliance with standard ethical practices in data handling. 

We advocate for the responsible use of ID-Booth by emphasizing its intended applications in privacy-preserving synthetic dataset generation rather than identity spoofing or deceptive practices. Future work should focus on bias mitigation strategies, ensuring fair representation across demographics, and improving safeguards against potential misuse. Further exploration of automated detection methods for synthetic faces could also enhance the transparency and accountability of generative models.

{\small
\bibliographystyle{ieee}
\bibliography{bibliography}
}


\clearpage
\setcounter{page}{1}



\section*{Supplementary material}


\subsection{Additional results and ablation studies}

\begin{figure}[tb!]
    \centering
    \includegraphics[width=\linewidth]{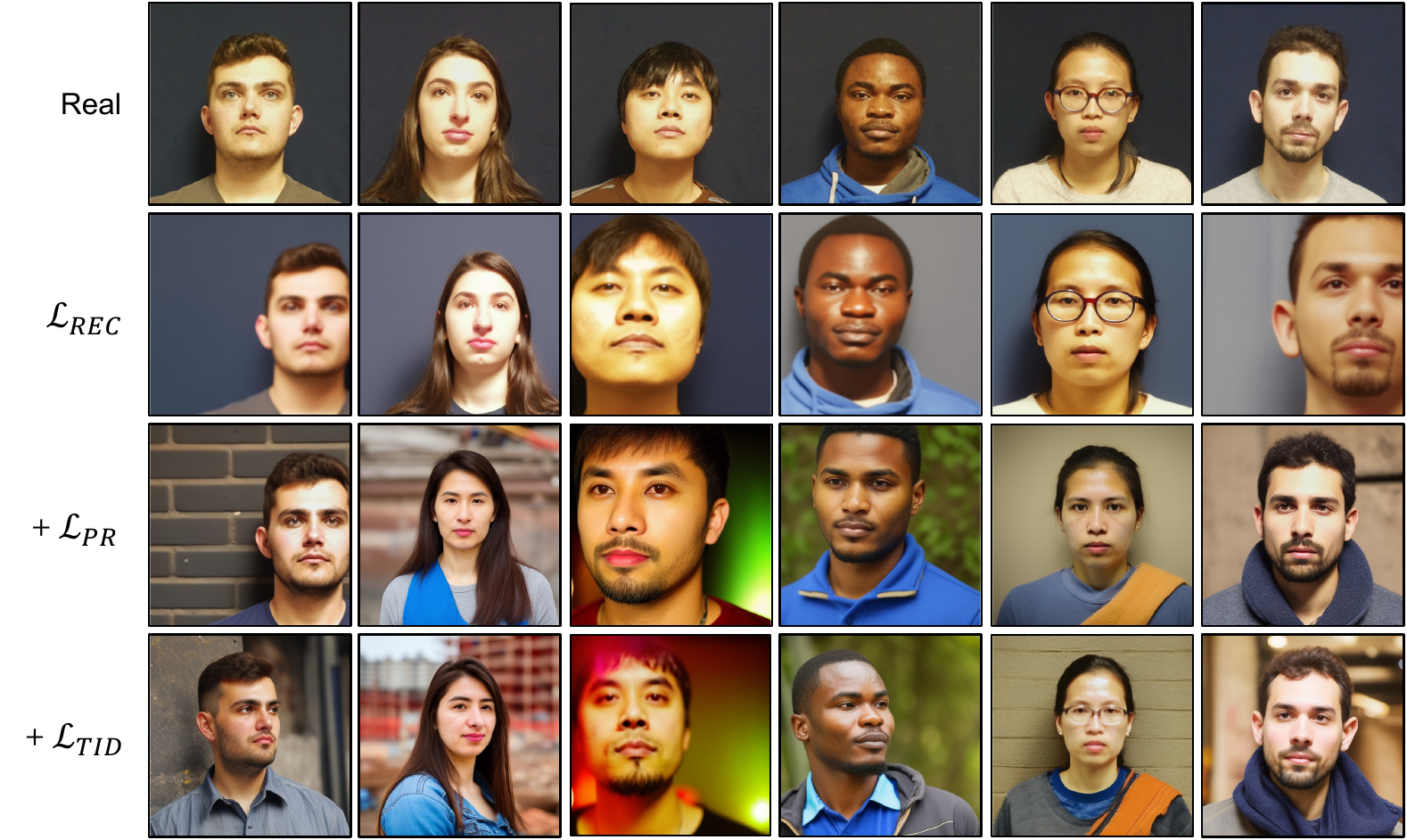}
    \vspace{-6mm}\caption{\textbf{Ablation study of ID-Booth training objectives.} Shown are sample images generated by the ID-Booth framework trained with different training objectives. Training only with $\mathcal{L}_{REC}$ generates images similar to the training set, disregarding the given prompts.  $\mathcal{L}_{PR}$ improves the diversity of samples but lowers identity consistency. Our proposed $\mathcal{L}_{TID}$ presents an objective that improves both diversity and identity consistency.  
    } 
    \label{fig:samples_loss_ablation}
    \vspace{-6mm}
\end{figure}

\noindent\textbf{Ablation of training objectives.}\label{sec:ablation_loss}
To determine the suitability of the selected configuration for our framework we first perform an ablation study regarding the training objectives utilized for fine-tuning the diffusion model. Following the evaluation methodology in Section~\ref{sec:experiments_and_results}, we report in Table~\ref{tab:quality_fidelity_diversity_ablation} results related to quality, fidelity and diversity on synthetic datasets consisting of $100$ samples per identity. In addition, we provide qualitative samples in Figure~\ref{fig:samples_loss_ablation} that demonstrate the effects of the different loss functions. As can be discerned from the samples, fine-tuning the Stable Diffusion 2.1 (SD-2.1)~\cite{rombach2022high_latent_diffusion} model with only $\mathcal{L}_{REC}$ leads to generated samples that closely resemble training samples from the Tufts Face Database (TFD)~\cite{TUFTS_panetta2018comprehensive}. Despite the utilized prompts, which specify the environment and subject pose as described in Section~\ref{sec:experiments_and_results}, the generated images often contain a uniform background, similar to the backdrop used in TFD~\cite{TUFTS_panetta2018comprehensive}. These observations are supported by results in Table~\ref{tab:quality_fidelity_diversity_ablation}, where we can observe that the synthetic images are more similar to images of FFHQ~\cite{stylegan_1_karras2019style} than the real-world images of TFD~\cite{TUFTS_panetta2018comprehensive} across all reported metrics. However, many of the images also contain minor artifacts including blurry faces, as also observed through considerably higher standard deviation of CR-FIQA~\cite{boutros2023cr_fiqa} scores. The density and coverage~\cite{naeem2020reliable} scores, which measure fidelity and diversity, are also rather low, especially when considering entire images. In combination with poor prompt adherence, these signs point to overfitting on characteristics of training samples from the Tufts Face Database (TFD)~\cite{TUFTS_panetta2018comprehensive}. 

The addition of the prior preservation objective $\mathcal{L}_{PR}$ resolves some of these issues by utilizing additional training images, which are generated by the pretrained model before fine-tuning. This prevents overfitting on undesired image characteristics of TFD~\cite{TUFTS_panetta2018comprehensive} (e.g., the background), thus facilitating better prompt adherence and in turn the generation of more diverse image samples, as depicted in Figure~\ref{fig:samples_loss_ablation}. While this is not reflected in quality-based metrics, we do see a noticeable improvement in terms of density (i.e., fidelity)~\cite{naeem2020reliable} on entire images, which point to backgrounds that resemble in-the-wild images. The face regions also become sharper, as denoted by improved CR-FIQA~\cite{boutros2023cr_fiqa} scores. However, these improvements come at the price of identity consistency both among synthetic and between synthetic and real samples, as discerned from verification measures of genuine and imposter distributions reported in Table~\ref{tab:ablation_pyeer_benchmarks_vs_real_and_vs_synthetic}.  

To address the identity-based issues raised by the higher diversity of samples enabled by $\mathcal{L}_{PR}$, we propose to utilize a triplet identity learning objective $\mathcal{L}_{TID}$. As seen by results in Table~\ref{tab:ablation_pyeer_benchmarks_vs_real_and_vs_synthetic}, the proposed identity-based objective achieves better results across the majority of measures, noticeably improving both intra-identity consistency and inter-identity separability. At the same time, the $\mathcal{L}_{TID}$ also ensures the generation of images that better match in-the-wild images of FFHQ~\cite{stylegan_1_karras2019style}, in terms of quality and diversity, as revealed by improvements in Fréchet Distance~\cite{heusel2017gans_FID}, Kernel Distance~\cite{binkowski2018demystifying_kernel_distance}, Coverage~\cite{naeem2020reliable} and per-class Vendi score~\cite{friedman_2024_vendi} in Table~\ref{tab:quality_fidelity_diversity_ablation}. Overall, the combination of the three objectives ensures the generation of diverse high-quality images of desired identities.





\begin{table*}[t!] 
\caption{\textbf{Ablation study of different training objectives and prompt components of ID-Booth through quantitative evaluation of quality, fidelity and diversity.} Quality is assessed with Fréchet Distance~\cite{heusel2017gans_FID} and Kernel Distance~\cite{binkowski2018demystifying_kernel_distance}, while fidelity and diversity are measured through Density and Coverage~\cite{naeem2020reliable}. Results are computed by comparing distributions of features extracted with DINOv2-ViT-L/14~\cite{oquab_2024_dinov2} from synthetic images and real-world images of FFHQ~\cite{stylegan_1_karras2019style}, considering either entire images or only the face region. Vendi score~\cite{friedman_2024_vendi} is used to evaluate intra-identity diversity, while CR-FIQA~\cite{boutros2023cr_fiqa} measures each synthetic sample separately, both without a reference dataset. 
\vspace{-2mm}
}
\centering
\resizebox{\linewidth}{!}{%
\begin{tabular}{llccccccccccc}
\toprule
 &  & \multicolumn{2}{c}{\bf{Fréchet Distance} $\downarrow$} &   \multicolumn{2}{c}{\bf{Kernel Distance} $\downarrow$}  &  \multicolumn{2}{c}{\bf{Density} $\uparrow$}  & \multicolumn{2}{c}{\bf{Coverage} $\uparrow$} & \multicolumn{2}{c}{\bf{Vendi score per ID $\uparrow$}}  & \bf{CR-FIQA}   \\ 
\bf{Data from} & \bf{Loss / Prompt}  &  Entire &  Face  &  Entire &  Face &  Entire &  Face & Entire & Face & Entire & Face & Face region  \\ 
\midrule  
TFD~\cite{TUFTS_panetta2018comprehensive} & $-$ & $2035.615$ & $1679.317$ &  $7.056$ & $5.779$ & $0.195$ & $0.623$ & $0.043$ & $0.120$ & $2.536$ & $3.132$  & $2.131 \pm 0.094$ \\
FFHQ~\cite{stylegan_1_karras2019style} & $-$ & $38.703$ & $33.799$ & $0.001$ & $0.001$ & $1.028$ & $1.007$ & $0.972$ & $0.970$ & $-$ & $-$ & $2.090 \pm 0.134$ \\ 

\midrule

\multirow{3}{*}{ID-Booth} & $\mathcal{L}_{REC}$ & $1256.490$ & $1258.637$ & $3.187$ & $3.657$ & $0.330$ & $0.551$ &  $0.119$ & $0.130$ & $9.754$ & $8.462$ & $2.079 \pm 0.226$ \\

 & + $\mathcal{L}_{PR}$ & $1374.696$ & $1371.129$ & $4.134$ & $4.484$ & \bm{$0.698$} & \bm{$0.575$} & $0.128$ & $0.131$ & $7.264$ & $6.705$ & ${2.187} \pm 0.134$ \\
& + $\mathcal{L}_{TID}$ & \bm{$1144.651$} & \bm{$1159.537$} & \bm{$2.778$} & \bm{$3.346$} & $0.536$ & $0.502$ & \bm{$0.157$} & \bm{$0.166$} & \bm{$13.510$} & \bm{$10.430$} & $2.143 \pm {0.181}$ \\

\midrule

\multirow{5}{*}{ID-Booth}  & Base prompt & $1849.488$ & $1889.947$ & $7.418$ & $7.639$ & $0.348$ & $0.310$ & $0.034$ & $0.031$ & $3.674$ & $3.185$ & $2.136 \pm 0.107$ \\

 & + \texttt{[B]} & $1394.284$ & $1454.935$ & $3.923$ & $4.631$ & $0.392$ & $0.362$ & $0.096$ & $0.097$ & $13.534$ & $11.761$ & $2.108 \pm 0.175$ \\
 
 & + Negative prompt & $1201.857$ & $1250.269$ & $3.261$ & $4.016$ & $0.543$ & $0.475$ & $0.151$ & $0.141$ & \bm{$14.876$} & \bm{$12.173$} & $2.158 \pm 0.159$ \\
& + \texttt{[G]} & $1185.778$ & $1230.030$ & $3.359$ & $4.085$ & \bm{$0.603$} & \bm{$0.531$} & $0.153$ & $0.153$ & $11.794$ & $9.358$ & $2.176 \pm 0.132$ \\
& + \texttt{[P]} & \bm{$1144.651$} & \bm{$1159.537$} & \bm{$2.778$} & \bm{$3.346$} & $0.536$ & $0.502$ & \bm{$0.157$} & \bm{$0.166$} & $13.510$ & $10.430$ & $2.143 \pm {0.181}$ \\

\bottomrule

\multicolumn{13}{l}{($\downarrow$ / $\uparrow$) -- Lower / Higher is better; (\textbf{Bold}) -- Best result; (\underline{Underline}) -- Second best result} \\
\end{tabular}
} 
\vspace{-4mm}
\label{tab:quality_fidelity_diversity_ablation}
\end{table*}

\begin{figure}[tb!]
    \centering
    \includegraphics[width=\linewidth]{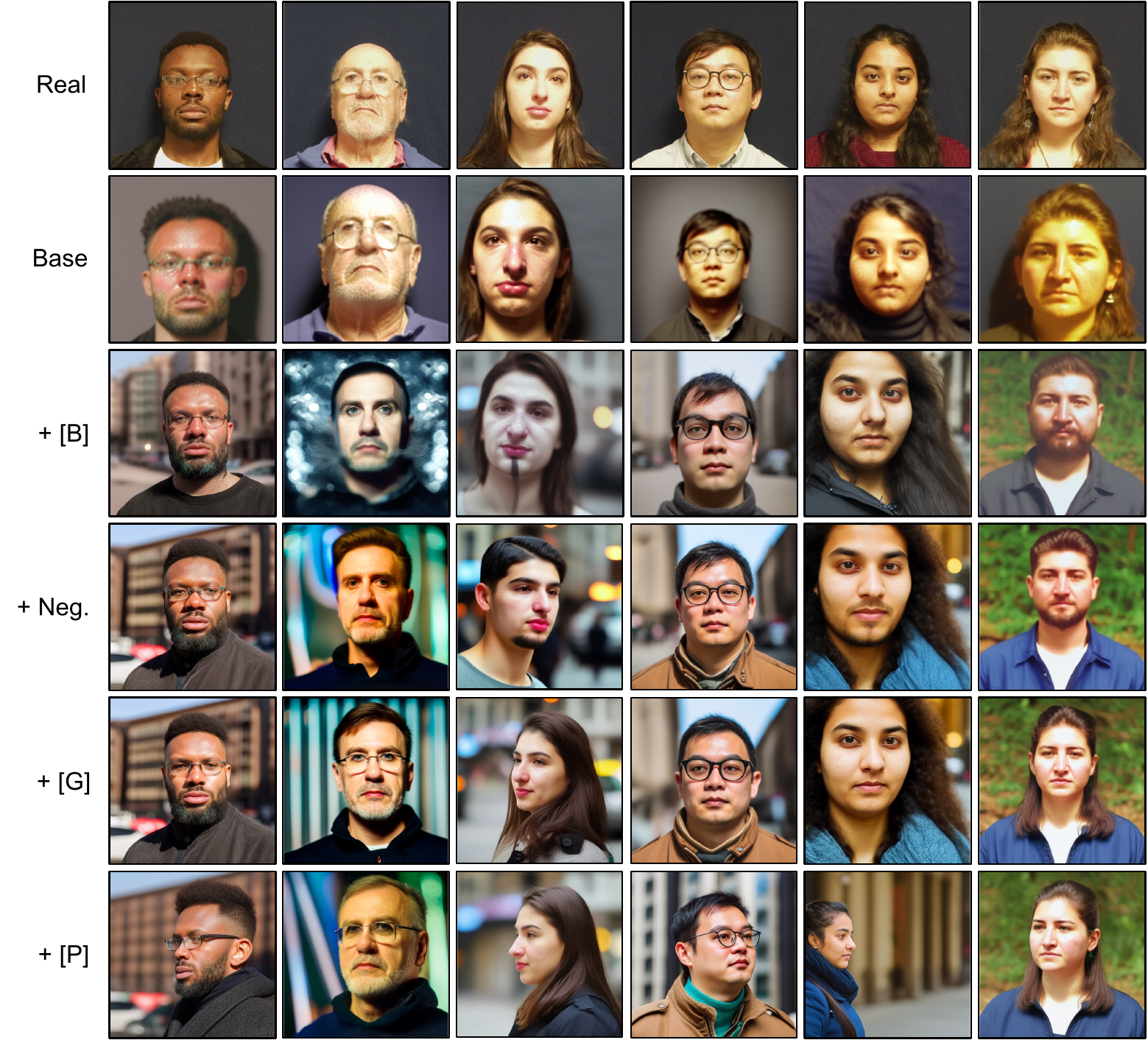}
    \caption{\textbf{Ablation study of prompt components used with ID-Booth.} Specifying the background \texttt{[B]} greatly improves diversity, but introduces artifacts. The negative prompt improves photorealism, while \texttt{[G]} solves issues related to gender. Adding the side-portrait component through \texttt{[P]} ensures the generation of more diverse poses. 
    } 
    \label{fig:sampless_ablation_prompts}
    \vspace{-6mm}
\end{figure}

\noindent\textbf{Ablation of inference prompts.}\label{sec:ablation_prompts}
As part of our experiments, we rely on a variety of prompts to facilitate the generation of diverse images that best match in-the-wild images of FFHQ~\cite{stylegan_1_karras2019style} while still retaining identity consistency. In this section, we investigate the effects of each prompt component presented in Section~\ref{sec:experiments_and_results}. This includes the background \verb|[B]|, the negative prompt, the subject gender \verb|[G]| and the subject portrait pose \verb|[P]|. We begin the experiments with the base prompt \verb|face portrait photo of [ID] person|, a more face-focused version of the training prompt, which had to be kept simple to ensure that the identity was correctly linked to the ID token during fine-tuning~\cite{ruiz2023dreambooth}. As can be observed from qualitative samples in Figure~\ref{fig:sampless_ablation_prompts}, generating images with solely the base prompt results in images that often portray subjects in a similar constrained environment as in the training images of TFD~\cite{TUFTS_panetta2018comprehensive}. The generated identities are consistent, as seen by verification results similar to TFD~\cite{TUFTS_panetta2018comprehensive} in Table~\ref{tab:ablation_pyeer_benchmarks_vs_real_and_vs_synthetic}. However, the quality, fidelity and diversity of images, reported in Table~\ref{tab:quality_fidelity_diversity_ablation}, is far from the desired characteristics of in-the-wild images of FFHQ~\cite{stylegan_1_karras2019style}, despite being closer than constrained images of TFD~\cite{TUFTS_panetta2018comprehensive}. The addition of the background prompt component \verb|[B]| drastically improves scores related to quality, fidelity and diversity of generated data in Table~\ref{tab:quality_fidelity_diversity_ablation}. The intra-identity diversity of samples is also notably improved, both in terms of the Vendi score~\cite{friedman_2024_vendi} as well as the standard deviation of head poses, reported in Table~\ref{tab:pose_estimation_ablation}. However, as can be discerned from samples in Figure~\ref{fig:sampless_ablation_prompts}, the additional complexity of the prompt also introduces prominent artifacts ranging from  unnatural backgrounds and face features to gender changes. In turn, this also immensely lowers intra-identity consistency and inter-identity separability, as reported in Table~\ref{tab:ablation_pyeer_benchmarks_vs_real_and_vs_synthetic}.

The use of an additional negative prompt, that guides the generation process away from undesired styles addresses some of these issues, as seen by improvements across all measures in Table~\ref{tab:quality_fidelity_diversity_ablation}. Furthermore, the negative prompt actually improves intra-identity diversity measured by the Vendi score~\cite{friedman_2024_vendi} as well as identity consistency slightly. 
Differently, specifying the gender with \verb|[G]|  of the subject drastically influences and improves identity consistency, as reported in Table~\ref{tab:ablation_pyeer_benchmarks_vs_real_and_vs_synthetic} and fixes the unintentional gender changes as observed in samples of Figure~\ref{fig:sampless_ablation_prompts}, which is also reflected in lower intra-identity diversity. 
Lastly, to also address the lack of pose diversity in the samples, we utilize the final prompt component \verb|[P]|, which specifies whether the image should be a portrait or a side-portrait. With this small change we can greatly influence the intra-identity diversity of head poses, as seen by drastic improvements in Table~\ref{tab:pose_estimation_ablation} in terms of the pitch, yaw and roll. In turn, this also improves results related to overall image diversity in Table~\ref{tab:quality_fidelity_diversity_ablation}. This addition also slightly negatively impacts the identity consistency, however, this is to be expected, which is to be expected as the features used for computing identity similarity are often affected by the head pose. Overall, the final constructed prompt ensures the generation of diverse high-quality images that better match the desired characteristics of in-the-wild images, while still ensuring identity consistency.


\begin{table}[t!] 
\caption{\textbf{Ablation of ID-Booth prompt components through pose estimation.}  Reported are the mean and standard deviation of standard deviation values of pitch, yaw and roll measured across samples of each identity with the 6DRepNet~\cite{hempel2024toward} head pose estimator.
\vspace{-2mm}
}
\centering
\resizebox{\linewidth}{!}{%
\begin{tabular}{llccc}
\toprule
 &  &   \multicolumn{3}{c}{\bf{Pose estimation }}    \\ 
       \bf{Data from} & \bf{Prompt} &  \bf{Pitch ($\sigma$ per ID)}  & \bf{Yaw ($\sigma$ per ID)}  & \bf{Roll ($\sigma$ per ID)} \\ 
    \midrule  
TFD~\cite{TUFTS_panetta2018comprehensive} & $-$ & $2.015 \pm 0.718$  & $26.297 \pm 4.609$  & $2.285 \pm 1.114$ \\
\midrule
\multirow{5}{*}{{ID-Booth}} & Base prompt   & $2.738 \pm 0.681$  & $7.254 \pm 3.470$  & $1.269 \pm 0.448$ \\
& + \texttt{[B]} & $5.147 \pm 0.962$  & $15.592 \pm 4.307$  & $2.522 \pm 0.949$ \\
& + Negative   & $4.787 \pm 0.925$  & $16.096 \pm 4.135$  & $2.370 \pm 0.892$ \\
& + \texttt{[G]} & $4.414 \pm 0.862$  & $17.114 \pm 4.504$  & $2.360 \pm 0.913$ \\
& + \texttt{[P]} & \bm{$6.641 \pm 2.662$}  & \bm{$33.527 \pm 7.569$}  & \bm{$5.637 \pm 2.920$} \\ 
\bottomrule

\multicolumn{5}{l}{($\downarrow$ / $\uparrow$) -- Lower / Higher is better; (\textbf{Bold}) -- Best result; (\underline{Underline}) -- Second best result} \\
\end{tabular}
} 
\vspace{-6mm}
\label{tab:pose_estimation_ablation}
\end{table}

\begin{table*}[tb!] 
\caption{\textbf{Ablation study of training objectives and prompt components of ID-Booth in terms of identity consistency and separability.}
Reported are verification measures of genuine and imposter distributions among synthetic or between synthetic and real images, constructed based on the cosine similarity of identity features obtained with a pretrained ArcFace-based recognition model~\cite{deng2019arcface}. Results of $\mathcal{L}_{REC}$ or the base prompt are not included when determining the best results as they produce samples that too closely match the real-world training samples of  TFD~\cite{TUFTS_panetta2018comprehensive}.
\vspace{-2mm}
}
\centering
\resizebox{\linewidth}{!}{%
\begin{tabular}{llcccccc}
\toprule
    \bf{Data setting}  & \bf{Loss / Prompt} & \bf{EER $\downarrow$} & \bf{FMR100 / 1000 $\downarrow$} & \bf{FNMR100 / 1000 $\downarrow$}  & \bf{Imposter $\mu \pm \sigma$ $\downarrow$} &  \bf{Genuine $\mu \pm \sigma$ $\uparrow$} & \bf{FDR $\uparrow$} \\ 
    \midrule
among TFD~\cite{TUFTS_panetta2018comprehensive} & Real data  & $0.002$ & $0.002$ / $0.002$ &  $0.001$ / $0.003$  & $0.021 \pm 0.0725$ & $0.871 \pm 0.070$ & $70.969$ \\
\midrule 

\multirow{3}{*}{among ID-Booth} & $\mathcal{L}_{REC}$ & $0.027$ & $0.042$ / $0.096$ & $0.226$ / $0.885$ & $0.056 \pm 0.079$ & $0.534 \pm 0.141$ & $8.704$ \\
\cmidrule{2-8}
& + $\mathcal{L}_{PR}$ 
& \underline{$0.055$} & \underline{$0.153$} / \underline{$0.337$} & \underline{$0.297$} / \underline{$0.919$} & \underline{$0.103 \pm 0.093$} & \bm{$0.499 \pm 0.141$} & \underline{$5.509$} \\

& + $\mathcal{L}_{TID}$ 
& \bm{$0.042$} & \bm{$0.095$} / \bm{$0.217$} & \bm{$0.249$} / \bm{$0.896$} & \bm{$0.059 \pm 0.082$} & \underline{$0.486 \pm 0.153$} & \bm{$6.073$} \\
\midrule 

\multirow{3}{*}{ID-Booth vs. TFD} & $\mathcal{L}_{REC}$ & $0.008$ & $0.007$ / $0.018$  & $0.003$ / $0.220$  & $0.021 \pm 0.073$ & $0.557 \pm 0.109$ & $16.709$ \\
\cmidrule{2-8}
& + $\mathcal{L}_{PR}$ & \underline{$0.046$} & \underline{$0.087$} / \underline{$0.184$} & \underline{$0.286$} / \bm{$0.684$} & \underline{$0.019 \pm 0.072$} & \underline{$0.406 \pm 0.155$} & \underline{$5.132$} \\ 
& + $\mathcal{L}_{TID}$ & \bm{$0.027$} & \bm{$0.044$} / \bm{$0.091$} & \bm{$0.110$} / \underline{$0.838$} & \bm{$0.017 \pm 0.072$} & \bm{$0.465 \pm 0.148$} & \bm{$7.402$} \\

\midrule
\multirow{5}{*}{among ID-Booth} & Base prompt & $0.002$ & $0.001$ / $0.002$  & $0.000$ / $0.009$  & $0.062 \pm 0.079$ & $0.799 \pm 0.100$ & $33.634$ \\
\cmidrule{2-8}
 & + \texttt{[B]} & $0.068$ & $0.143$ / $0.244$ & $0.561$ / $0.942$ & \underline{$0.060 \pm 0.080$} & $0.454 \pm 0.174$ & $4.238$ \\
 & + Negative prompt & $0.064$ & $0.137$ / $0.265$  & $0.520$ / $0.932$  & $0.064 \pm 0.082$ & $0.485 \pm 0.179$ & $4.577$ \\
 & + \texttt{[G]} & \bm{$0.036$} & \bm{$0.069$} / \bm{$0.147$}   & \bm{$0.198$} / \bm{$0.799$}  & $0.070 \pm 0.087$ & \bm{$0.533 \pm 0.149$} & \bm{$7.210$} \\
& + \texttt{[P]} & \underline{$0.042$} & \underline{$0.095$} / \underline{$0.217$} & \underline{$0.249$} / \underline{$0.896$} & \bm{$0.059 \pm 0.082$} & \underline{$0.486 \pm 0.153$} & \underline{$6.073$} \\
\midrule 
\multirow{5}{*}{ID-Booth vs. TFD} & Base prompt & $0.004$ & $0.003$ / $0.007$ & $0.000$ / $0.118$  & $0.018 \pm 0.073$ & $0.662 \pm 0.095$ & $28.963$ \\
\cmidrule{2-8}
 & + \texttt{[B]} & $0.038$ & $0.074$ / $0.128$  & $0.274$ / $0.744$   & $0.019 \pm 0.074$ & \bm{$0.462 \pm 0.160$} & $6.307$ \\
 & + Negative prompt & \underline{$0.036$} & \underline{$0.067$} / \underline{$0.142$}   & \underline{$0.238$} / \underline{$0.673$}   & \bm{$0.014 \pm 0.075$} & {$0.466 \pm 0.163$} & \underline{$6.389$} \\
 & + \texttt{[G]} & \bm{$0.021$} & \bm{$0.030$} / \bm{$0.067$}   & \bm{$0.067$} / \bm{$0.571$}  & \underline{$0.017 \pm 0.073$} & \underline{$0.478 \pm 0.138$} & \bm{$8.736$} \\
& + \texttt{[P]} & {$0.042$} &  {$0.095$} / {$0.217$} & {$0.249$} / {$0.896$} & {$0.059 \pm 0.082$} & {$0.486 \pm 0.153$} & {$6.073$} \\
 
\bottomrule  
\multicolumn{6}{l}{($\downarrow$ / $\uparrow$) -- Lower / Higher is better; (\textbf{Bold}) -- Best result; (\underline{Underline}) -- Second best result}
\end{tabular}
} 
\vspace{-4mm}

\label{tab:ablation_pyeer_benchmarks_vs_real_and_vs_synthetic}
\end{table*}

\vspace{0.8mm}\noindent\textbf{Real-world time requirements.}
Fine-tuning with our proposed ID-Booth framework takes $31$ minutes on average for each identity of TFD~\cite{TUFTS_panetta2018comprehensive} with an Nvidia A100 GPU. Similarly, fine-tuning with the PortraitBooth-based~\cite{peng2024portraitbooth} approach requires $30$ minutes on average, likely due to the simpler training objective. Differently, DreamBooth~\cite{ruiz2023dreambooth}, which does not utilize an identity-based objective, requires only $20$ minutes for fine-tuning on average, as it does not require the decoding of latent samples during training.

\subsection{Details regarding the evaluation methodology}

\noindent\textbf{Measuring image quality, fidelity and diversity.}
To evaluate the synthesis capabilities of the fine-tuned models we compare the produced synthetic images with in-the-wild face images of FFHQ~\cite{stylegan_1_karras2019style}.
To this end, we utilize the following performance measures:

\begin{itemize}
    \item \textbf{Fréchet Distance~\cite{heusel2017gans_FID}} which estimates the overall quality of synthetic images, by evaluating the difference between distributions of image features. For this, the original implementation (i.e., FID) utilizes an Inception-v3 network~\cite{szegedy2016rethinking_inception} pretrained on ImageNet~\cite{deng2009imagenet} as a feature extractor. Differently, we rely on features extracted with DINOv2-ViT-L/14~\cite{oquab_2024_dinov2}, which offers a representation space that is more consistent with human evaluators and thus better suited for evaluating generative approaches~\cite{stein2023exposing_dgm_eval}.
    
    \item \textbf{Kernel distance~\cite{binkowski2018demystifying_kernel_distance}} represents an alternative to Fréchet Distance by utilizing the maximum mean discrepancy to measure the distance between distributions. Throughout our experiments we rely on the third degree polynomial kernel, following existing works~\cite{stein2023exposing_dgm_eval} and compute the distance on image features of DINOv2-ViT-L/14~\cite{oquab_2024_dinov2}.

    \item \textbf{Density and Coverage~\cite{naeem2020reliable}}, which measure the fidelity and diversity, respectively, by considering the distance between nearest neighbour embeddings of images. Density and coverage address issues with previous methods, namely precision and recall~\cite{kynkaanniemi2019improved_precision_recall}, by providing a measure that is more resilient to outliers. While the original implementation relied on features extracted with the VGG16 network~\cite{simonyan2014very_VGG} pretrained on ImageNet~\cite{deng2009imagenet}, we instead utilize features of DINOv2-ViT-L/14~\cite{oquab_2024_dinov2}.

    \item \textbf{Vendi score~\cite{friedman_2024_vendi}}, which measures dataset diversity without the need for a reference dataset. However, when conditioned on the class (i.e., the identity) it can also be used to quantify intra-class diversity of samples. It can therefore be used alongside Coverage~\cite{naeem2020reliable} to gain better insight into the diversity of generated data. To compute it we rely on image features extracted with DINOv2-ViT-L/14~\cite{oquab_2024_dinov2}.
    
    \item  \textbf{Certainty Ratio Face Image Quality Assessment (CR-FIQA)~\cite{boutros2023cr_fiqa}} measure, which is designed specifically for evaluating the quality of face images. It measures the  quality through the relative classifiability of a given face image with a pretrained ResNet-101 network~\cite{he2016deep_resnet}. 
    
\end{itemize}

Here, it should be noted that the fine-tuned diffusion models produce images that often contain more context than just the face region, differently from the FFHQ dataset~\cite{stylegan_1_karras2019style}. Thus, to a allow for a fair evaluation of specifically the face region we preprocess the generated images, following the preprocessing steps of FFHQ~\cite{stylegan_1_karras2019style}. This includes first detecting facial landmarks with the Multi-Task Cascaded Convolutional Neural Network (MTCNN)~\cite{zhang2016MTCNN} and then defining an affine transform to align them to a set of predefined positions. Finally, images are cropped to a resolution of $112\times112$, suitable for the AdaFace-based recognition model~\cite{kim2022adaface}. 
For Fréchet Distance~\cite{heusel2017gans_FID}, Kernel Distance~\cite{binkowski2018demystifying_kernel_distance}, as well as density and coverage~\cite{naeem2020reliable}, which utilize both synthetic and real-world distributions for evaluation, we utilize the entire synthetic datasets with $100$ samples per identity and $10,000$ samples from FFHQ~\cite{stylegan_1_karras2019style}. 
To obtain a baseline for the scores, we also compare samples from the Tufts Face Database~\cite{TUFTS_panetta2018comprehensive} with samples from FFHQ~\cite{stylegan_1_karras2019style}, as well as compare two sets of $10,000$ samples from FFHQ~\cite{stylegan_1_karras2019style} between each other.

To further analyze intra-identity diversity in a more explainable manner, we also investigate the pitch, yaw and roll of samples for each identity with the state-of-the-art \textbf{6DRepNet~\cite{hempel2024toward}} head pose estimator, which is pretrained on the 300W-LP dataset~\cite{zhu2016face}.

\vspace{0.8mm}\noindent\textbf{Assesment of identity consistency and separability.} We also investigate the generated images in terms of the identity aspect in order to better understand the consistency and separability of identities of generated datasets. For this purpose we utilize genuine and imposter score distribution plots, based on the cosine similarity of features extracted with the pretrained ArcFace recognition model~\cite{deng2019arcface}. Below we provide descriptions of the verification measures that we utilize throughout the experiments: 
\begin{itemize}
    \item \textbf{Equal Error Rate (EER)}~\cite{maio2002fvc2000_EER}, which  is the point on the Receiver Operating Characteristics (ROC) curve, where the False Match Rate (FMR) equals the False Non-Match Rate (FNMR).
    
    \item \textbf{FMR100} and \textbf{FMR1000}, which report the lowest the False Non-Match Rate (FNMR)  achieved at a False Match Rate (FMR) of $1.0\%$ or $0.1\%$ respectively. 

    \item \textbf{FNMR100} and \textbf{FNMR1000} that represent the lowest the False Match Rate (FMR)  achieved at a False Non-Match Rate (FNMR) of $1.0\%$ or $0.1\%$ respectively. 

    \item \textbf{Fisher Discriminant Ratio~(FDR)}~\cite{poh2004_FDR}, which quantifies the separability of genuine and imposter distributions.

\end{itemize}

\vspace{0.8mm}\noindent\textbf{Recognition experiments.}
In the experiments, we train the AdaFace recognition model~\cite{wang2018cosface} on the produced synthetic datasets, as described in Section~\ref{sec:details_recognition_experiments}. To determine their suitability, we evaluate the performance of the model  on five real-world verification benchmarks. These include: 
\begin{itemize}
    \item \textbf{Labeled Faces in the Wild (LFW)}~\cite{huang2008labeled_LFW}, which is an unconstrained web-scraped verification dataset of $13,233$ face images of $5749$ identities. 
    \item \textbf{Cross-Age Labeled Faces in the Wild (CA-LFW)}~\cite{zheng2017cross_CALFW}, which is a subset of  LFW~\cite{huang2008labeled_LFW} with $7156$ images of $2996$ identities, aimed at evaluating verification performance across a given age gap. 

    \item \textbf{Cross-Pose Labeled Faces in the Wild (CP-LFW)}~\cite{zheng2018cross_CPLFW}, which is a LFW~\cite{huang2008labeled_LFW} subset that is suited specifically for evaluating cross-pose verification performance. It includes $5984$ face images of $2296$ identities captured in various poses.

    \item \textbf{AgeDB-30}~\cite{moschoglou2017agedb}, which is a dataset of in-the-wild face images, suited for evaluating verification performance across a $30$ year age gap. The dataset comprises $16,488$ images of $568$ identities. 
    
    \item \textbf{Celebrities in Frontal-Profile in the Wild (CFP-FP)}~\cite{sengupta2016frontal_CFPFP}, which is a verification dataset that is aimed at evaluating cross-pose performance, in particular of frontal and profile poses. In total, it contains $7000$ images of $500$ identities, each with $10$ frontal and $4$ profile images.
\end{itemize}
Each benchmark is formed with $3000$ genuine and $3000$ imposter image pairs of a given verification dataset, with an image resolution of $112 \times 112$. To limit the influence of race and gender, the CA and CP verification pairs are sampled from the same race and gender.

\end{document}